\documentclass[Afour,sageh,times]{sagej}

\usepackage{moreverb,url}
\usepackage{amsmath,amssymb,amsfonts}
\usepackage{algorithmic}
\usepackage{array}
\usepackage[caption=false,font=normalsize,labelfont=sf,textfont=sf]{subfig}
\usepackage{caption}
\usepackage{textcomp}
\usepackage{stfloats}
\usepackage{url}
\usepackage{pifont}
\newcommand{\CheckmarkBold}{\ding{51}} 
\newcommand{\XSolidBrush}{\ding{55}}   
\usepackage{graphicx}
\usepackage{balance}
\usepackage[table]{xcolor}  
\usepackage{colortbl}       

\definecolor{red}{rgb}{1.00,0.00,0.00}
\definecolor{blue}{rgb}{0.00,0.00,1.00}
\definecolor{green}{rgb}{0.30, 0.50,0.00}
\definecolor{lightgray}{gray}{0.9}
\definecolor{purple}{rgb}{0.60,0.10,0.90}
\newcommand{\cred}[1] {\textcolor{red}{#1}}
\newcommand{\cblue}[1] {\textcolor{blue}{#1}}

\usepackage{cite}

\usepackage{mathtools}
\usepackage{proof}

\usepackage{tabularx}
\usepackage{booktabs}
\usepackage{multicol}
\usepackage{multirow}
\usepackage{floatrow}
\usepackage{float}
\usepackage{url}
\usepackage{tikz}
\usepackage{ctable}
\usepackage{pgf}
\usetikzlibrary{shapes,external}
\usetikzlibrary{shapes.multipart}
\usetikzlibrary{calc, positioning, automata}

\usepackage[colorlinks,bookmarksopen,bookmarksnumbered,citecolor=red,urlcolor=red]{hyperref}

\newcommand\BibTeX{{\rmfamily B\kern-.05em \textsc{i\kern-.025em b}\kern-.08em
T\kern-.1667em\lower.7ex\hbox{E}\kern-.125emX}}

\def\volumeyear{2024}
\begin{document}

\runninghead{Smith and Wittkopf}

\title{Enhancing Interpretability and Interactivity in Robot Manipulation: A Neurosymbolic Approach}

\author{Georgios Tziafas$^{1}$ and Hamidreza Kasaei$^{1}$
\thanks{$^{1}$Department of Artificial Intelligence,
        University of Groningen, The Netherlands
        {\tt\small \{g.t.tziafas,hamidreza.kasaei\}@rug.nl}}}

\corrauth{Georgios Tziafas, \\ Interactive Robot Learning Lab, Department of Artificial Intelligence \\
Bernoulli Institute for Mathematics, Computer Science and Artificial Intelligence, University of Groningen \\
Groningen, the Netherlands,
Zernikepark 1, 9747 AA Groningen}

\email{g.t.tziafas@rug.nl}

\begin{abstract}
In this paper we present a neurosymbolic architecture for coupling language-guided visual reasoning with robot manipulation.
A non-expert human user can prompt the robot using unconstrained natural language, providing a referring expression (REF), a question (VQA), or a grasp action instruction. 
The system tackles all cases in a task-agnostic fashion through the utilization of a shared library of primitive skills.
Each primitive handles an independent sub-task, such as reasoning about visual attributes, spatial relation comprehension, logic and enumeration, as well as arm control.
A language parser maps the input query to an executable program composed of such primitives, depending on the context.
While some primitives are purely symbolic operations (e.g. counting), others are trainable neural functions (e.g. visual grounding), therefore marrying the interpretability and systematic generalization benefits of discrete symbolic approaches with the scalability and representational power of deep networks.
We generate a 3D vision-and-language synthetic dataset of tabletop scenes in a simulation environment to train our approach and perform extensive evaluations in both synthetic and real-world scenes.
Results showcase the benefits of our approach in terms of accuracy, sample-efficiency, and robustness to the user's vocabulary, while being transferable to real-world scenes with few-shot visual fine-tuning.
Finally, we integrate our method with a robot framework and demonstrate how it can serve as an interpretable solution for an interactive object-picking task, achieving an average success rate of 80.2\%, both in simulation and with a real robot. 
We make supplementary material available in \href{https://gtziafas.github.io/neurosymbolic-manipulation/}{https://gtziafas.github.io/neurosymbolic-manipulation}.
\end{abstract}

\keywords{Neurosymbolic Robotics, Natural Human-Robot Interaction}

\maketitle

\section{Introduction}
\begin{figure}[!h]
    \centering
    \includegraphics[width=\textwidth]{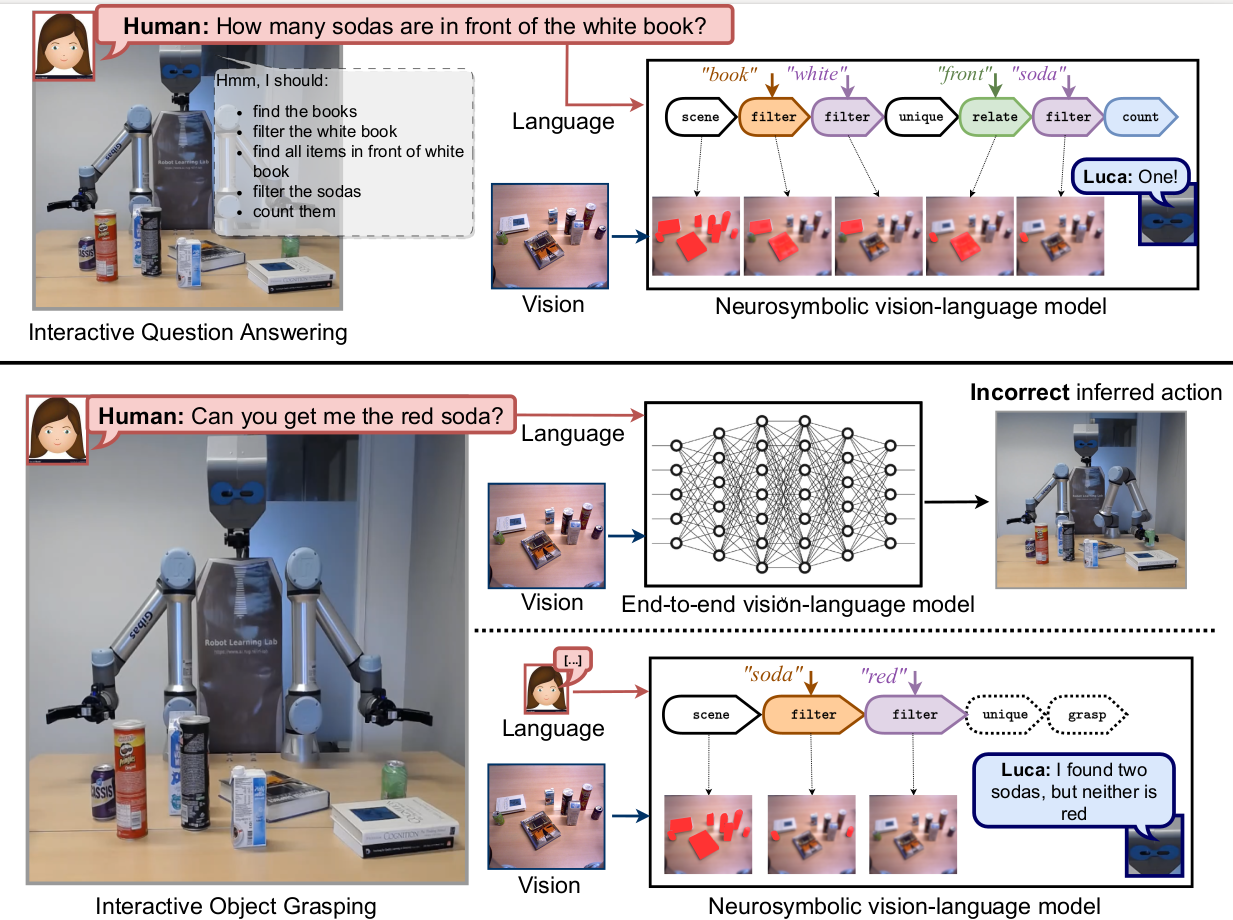}
    \caption{\footnotesize
    Example scenarios where a human user interacts with the robot in natural language. Understanding the
input question / instruction often requires reasoning about properties or relations of appearing objects in a compositional manner. 
Neurosymbolic approaches parse the input question into the underlying reasoning program and execute it step-by-step in order to reach the final answer (\textit{top}). 
Similarly, we propose a neurosymbolic model that represents grasp policies as programs in an interpretable formal language. End-to-end vision-language-grasping methods learn a policy directly from raw inputs and thus actions are generated regardless of the scene content. In the second example \textit{(bottom)}, there is no red soda for the robot to grasp, but only our approach is able to capture this and communicate it to the user.}
    \label{fig:Fig1}
\end{figure}

As modern developments in robotics are beginning to move robots from purely industrial to human-centric environments, it becomes essential for them to be able to interact naturally with humans. 
This necessity poses two additional challenges to traditional autonomy, as the agent is expected to be \textit{interactive}, i.e. able to receive task-specific instructions from its human cohabitants, as well as \textit{interpretable}, i.e. complete the task in a manner that is fully explainable to non-expert users.
The second feature is of particular interest, as it enables humans to diagnose and correct erroneous robot behaviors via online interaction, e.g. through free-form natural language.
Grounding perception and action in natural language has been a central theme in recent computer vision and robotics literature, from language-grounded 3D vision \citep{referit3d, scanrefer, scanqa}, to language-conditioned manipulation \citep{LanguageConditionedIL, Lynch2020LanguageCI, Lynch2020LanguageCI, BCZ}, to integrated language-based systems \citep{Socratic, saycan, InnerME} for high-level reasoning and task planning. 
Across domains, language has shown to be a great inductive bias for effective robot learning, however, methods still struggle with grounding fine-grained concepts beyond object category (i.e., visual attributes and spatial relations) \citep{CLIPort}, as well as reasoning about them in an algorithmic fashion (e.g. counting).
The end-to-end nature of most approaches leads to additional limitations, namely: a) \textit{lack of interpretability}, as the underlying reasoning process required to solve the task is captured \textit{implicitly} in the network's representations and thus cannot be retrieved from the output, b) \textit{data-hungriness}, i.e., need of large vision-language datasets that sufficiently sample the space of all possible concept combinations, and c) \textit{closed-endedness}, as the end-to-end policy is trained for a fixed agent/environment and catalog of concepts and tasks.

\begin{figure*}[!t]
    \centering
    \includegraphics[width=\textwidth]{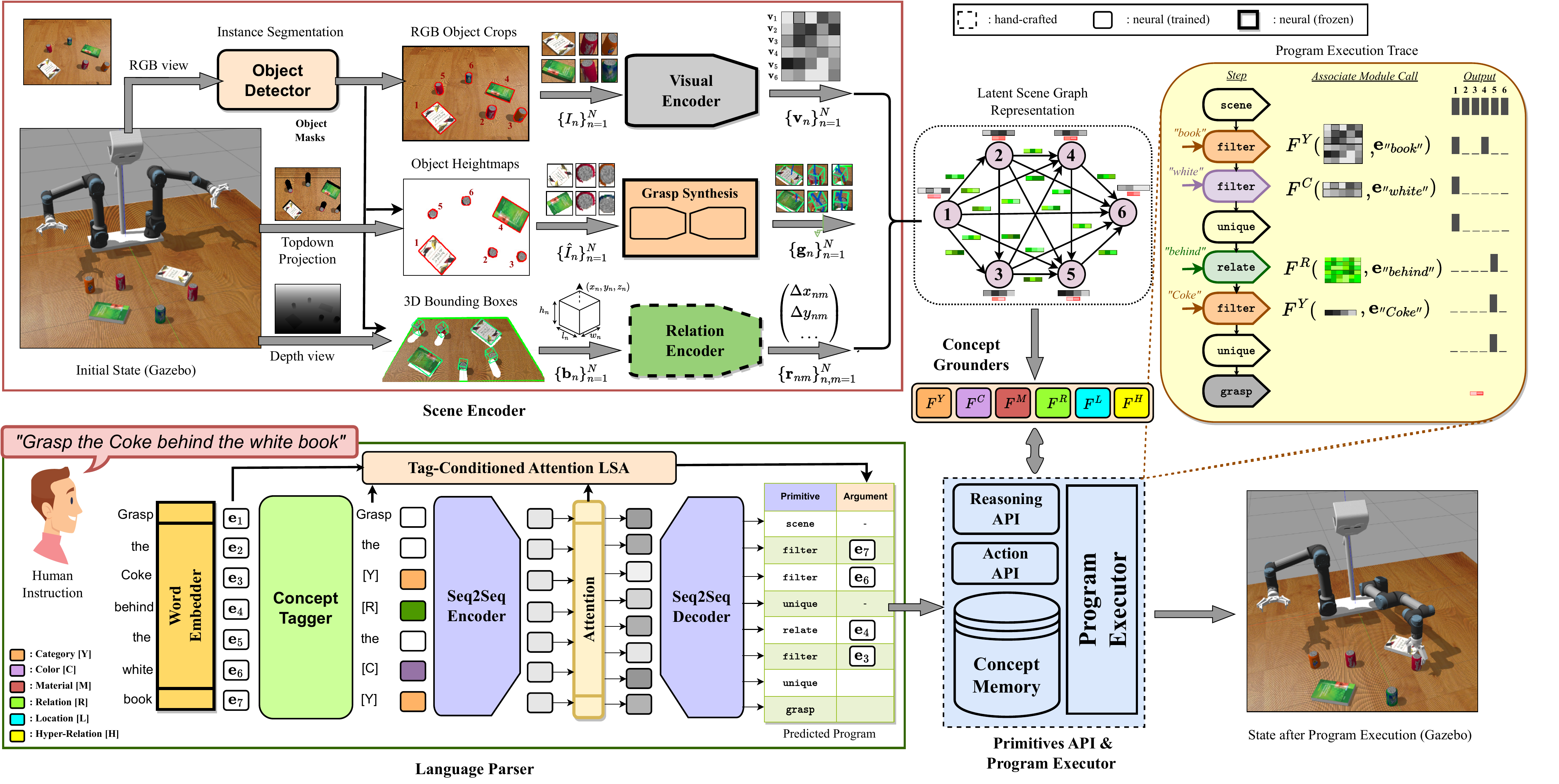}
    \caption{\footnotesize
    A schematic of the proposed framework. First, objects are segmented and localized in 3D space \textit{(top left)} and the scene is represented as a graph of extracted object-based features (visual, grasp pose) as nodes and spatial relation features as edges \textit{(top middle)}.
    A human user provides an instruction and a language parser generates an executable program \textit{(bottom left)}, built out of a primitives library \textit{(bottom middle)}.
    A program executor utilises a set of concept grounding modules to ground words to different objects \textit{(center)} and executes the predicted program step-by-step \textit{(top right)}, in order to identify the queried object and instructs the robot to grasp it \textit{(bottom right)}.}
    \label{fig:Fig2}
    \vspace{-3mm}
\end{figure*}

We believe that these limitations stem from the holistic fashion in which most methods couple language with perception. 
In particular, they either rely on visual-text feature fusion in a joint space (\citep{holisticNlgrasp, holisticNlgrasp1, INGRESS, LanguageConditionedIL,  CLIPort}), or FiLM-conditioning \citep{FILM} the visual network with a sentence-wide embedding of the language input (\citep{BCZ, saycan}).
We argue that this methodology fails to exploit the compositional nature of language, instead relying on variance to learn one-to-one correspondences between task descriptions and robot behavior.
For instance, consider a scenario like the one shown in Fig.~\ref{fig:Fig1}, where a human asks a question about the scene: (\textit{e.g. ``How many sodas are in front of the white book?"}). The task requires grounding multiple different concepts (i.e., visual - \textit{``book", ``white"}, spatial -  \textit{``front"} and symbolic - \textit{``How many"}) and reason about the intermediate results to reach a final answer.
Our intuition is that, for a human, the logic behind solving this task is compositional (a hierarchy of primitive steps) and disentangled from perception, meaning that the reasoning steps illustrated in Fig.~\ref{fig:Fig1} can be generalized to all similar questions regardless of the actual scene content.

Such intuition is encapsulated within neurosymbolic frameworks \citep{clevr, clevr-ref, ns-cl, ns-vqa}, that propose to further inject prior knowledge about language in the form of symbolic programs \citep{ns-vqa}, which \textit{explicitly} describe the underlying reasoning process.
The overall task is decomposed into independent sub-tasks (primitives), and each one is implemented as a symbolic module in a \textit{Domain-Specific Language (DSL)}.
The idea is to use deep neural nets as parsing tools - from images to structured object-based representations and from text queries to programs - and pair them with a symbolic engine for executing the parsed program in the scene representation to reach an answer.
By disentangling perception and language understanding (\textit{neural}) from reasoning (\textit{symbolic}), neurosymbolic systems address several of the highlighted limitations, i.e., other than a final answer, they output a formal interpretable representation of the underlying reasoning process (see Fig.~\ref{fig:Fig1}).
Furthermore, utilizing programs as a prior for learning grants the system highly sample-efficient and aids in generalization to unseen concept-task combinations \citep{ns-cl, ns-vqa}.
However, prior arts are limited to REF/VQA tasks, and associated datasets \citep{clevr, clevr-ref} model abstract synthetic domains with a poor variety of object and relation semantics.
Proposed methods also fix their DSL to be aware of the domain vocabulary (i.e., primitives are coupled with concept arguments), limiting them to the concepts encountered at training time.

In this work, we wish to propagate neurosymbolic reasoning to the robotics field and utilize it as an auxiliary process for interpretable robot manipulation. 
To that end, we generate a synthetic 3D vision-and-language dataset with a broad collection of object categories, attribute and relation concepts.
We design a corresponding DSL and re-formulate components of previous neurosymbolic recipes in order to handle the open-vocabulary requirement (see schematic in Fig.~\ref{fig:Fig2}).
In particular, we decompose the language-to-program module into two steps, first identifying concepts in the sentence to create an abstracted version of the query and then feeding it to a seq2seq network to generate the program, thus relieving the latter component from having to deal with the specific concept vocabulary of the training set.
To ground (potentially unseen) concepts in the image, we use concept grounding networks that operate on latent object-relation features, serving as an alternative to classification.
We compare our method with other holistic / neurosymbolic baselines in terms of accuracy and sample-efficiency and show that it can be transferred to real images via few-shot fine-tuning of the visual grounder network. 
We further integrate our model with a robot framework and test its performance in an interactive object grasping task, where we show that its highly interpretable nature allows us to study the distribution of failure modes across the different system components.
We close our evaluation by showing that the method can be efficiently extended to more manipulation tasks with the cost of a few hundred relevant instruction-program annotations.
In summary, the key contributions of this work are threefold: 
\begin{itemize}
    \item We generate a synthetic dataset of household objects in tabletop scenes for REF/VQA/grasping tasks, equipped with program annotations for reasoning, and collect a small-scale real-scene counterpart for evaluation. We make both datasets publicly available.
    
    \item We propose a neurosymbolic framework that integrates instance segmentation, visual / spatial grounding, semantic parsing and grasp synthesis in a vocabulary-agnostic formulation that supports application in unseen vocabulary, granting it transferable to novel concepts / tasks with minimal adaptation. 
    
    \item We perform extensive experiments to show the merits of our approach in terms of (\textit{i}) interpretable, highly accurate and sample-efficient reasoning, evaluated through a VQA task, (\textit{ii}) robustness to users vocabulary, (\textit{iii}) efficient adaption to natural scenes and more manipulation tasks and (\textit{iv}) applicability for interpretable interactive object grasping, tested both in simulation and with a real robot.  

\end{itemize}

\section{Related Works}
\label{sota}
\textbf{Grounding referring expressions} Grounding visual and spatial concepts expressed through language is a central challenge for an interactive robot.
Deep learning literature poses this through the task of grounding referring expressions (REF) \citep{refoco, flickr}, i.e., localizing an object in a scene from a natural language description.
Methods usually employ a two-stage detect-then-rank approach, leveraging off-the-shell detectors to first propose objects and then rank their object-query matching scores through CNN-LSTM feature fusion \citep{traditional1, traditional2, baselineMLP} or attention mechanisms \citep{baselineAttn}.
Alternatively, richer cross-modal contextualization between images and words is pursued through external syntactic parsers \citep{external1, extparser}, graph attention networks \citep{graph1, graph2, graph3} or Transformers \citep{visual-bert, vilbert, UNITER, ernie-vil}. 
Single-stage methods \citep{1stage, zsg, vgTRM} attempt to alleviate the object proposal bottleneck by densely fusing textual with scene-level visual features to create joint multimodal representations. 
Transferring from large-scale vision-language pretraining \citep{CLIP, AlignBF} aids in out-of-distribution generalization and can be used in zero-shot setups  \citep{ReCLIPAS} or for open-vocabulary object detection \citep{vild}.
REF has been also extended to the 3D domain \citep{referit3d, scanrefer}, where similar to 2D, most methods employ detect-then-rank pipelines, fusing textual features with segmented point-clouds \citep{referit3d, Zhao20213DVGTransformerRM} or RGB-D views \citep{Liu2021ReferitinRGBDAB, Huang2022MultiViewTF}.
All the above approaches follow the holistic methodology, hence as argued in the previous section, suffer from data-hungriness and lack the desired interpretability property.

More closely to our work, modular approaches \citep{CMN, Mattnet, nmtree} decompose the grounding task in independent modules (e.g. entities, attributes, relations) and predict their composition based on the query's structure with a language parser.
Such methods use soft attention-based parsers that are trained end-to-end with the rest of the modules using weak supervision.
In \citep{tziafas}, the modules are trained separately using dense attribute- and relation-level supervision from synthetic data and are linked to words using a tagger network.
However, module composition is handled by a linguistics-inspired heuristic, and hence, it is limited to referring expressions that follow a standard subject-relation-object syntax.
Similarly, we use a tagger and dense synthetic supervision to train our modules but replace the heuristic with a seq2seq network, that can map arbitrary syntactic structure into a formal representation (program), expressed via a DSL.
With this, we can extend the scope of the parser from grounding referring expressions to VQA and eventually robot action, by adding the associated modules in our DSL.

\textbf{Neurosymbolic reasoning} Early works in modular networks for VQA \citep{Johnson2017, Hu2017LearningTR, Hudson2018CompositionalAN, NMN, nmn2, nmn3} demonstrate capacities for compositional vision-language reasoning, by integrating independent modules instead of end-to-end learners.
More recently, a neurosymbolic model for VQA (NS-VQA) \citep{ns-vqa} in CLEVR \citep{clevr} and its extensions to natural images \citep{SoTA1, nsm, lggr} utilize a formal DSL and a symbolic program executor to run programs on parsed scene representations. 
Program generation and scene parsing (i.e. localization and attribute recognition) are trained separately and interface with the executor only at test-time.
In such works, however, the scene is represented as a table of attribute labels \citep{ns-vqa} or features \citep{ns-cl}, without any relation information.
Resolving spatial relations is then achieved by using concept-specific heuristics as primitives (e.g. relate left).
Visual attribute concepts are either classified \citep{ns-vqa} and coupled with primitives or matched with concept representations learned jointly from a closed-set \citep{ns-cl}. 
This formulation makes the system fixed to the concept vocabulary encountered during training.
In our work, we integrate relation concepts with object-based features in a latent scene graph representation and make our primitives vocabulary-agnostic, allowing extension to novel concepts without touching the DSL, via concept grounding networks.
Like NS-CL \citep{ns-cl}, we enable open-vocabulary parsing by replacing lexical items in the input query with their corresponding concepts.
Unlike NS-CL, which assumes access to ground truth tags, we learn the word-to-concept mapping through a tagging sub-module.

There are a few works that similar to our paper apply neurosymbolic reasoning in the robotics domain. ProgramPort \citep{ProgrammaticallyGC} uses a CCG parser to construct programs, CLIP~\citep{CLIP} for grounding attributes and learn a specialized pick-and-place module for selecting affordances on a top-down 2D image end-to-end.
In \citep{Kalithasan2022LearningNP}, the authors use a similar neural scene encoder and semantic parser to our framework, but focus on learning transition models that can predict future states of objects for planning.
Similarly, PDSketch~\citep{Mao2023PDSketchIP} defines a DSL that allows human to draw program sketches for specific tasks, and learn ellaborate transition models that include continuous parameters for actions. Our work differentiates itself by introducing the latent scene graph representation, which already contains action-related parameters (i.e. grasp poses) and focuses on generalizing semantic parsing and reasoning.


\textbf{Language-guided manipulation} In the robotics field, language-conditioning has been an emergent theme in RL-based \citep{lRL1, RLsurvey} and IL-based \citep{LanguageConditionedIL, Lynch2020LanguageCI, Lynch2020LanguageCI, BCZ} manipulation.
Such methods require prohibitive training resources or several hours of human teleoperation data, dedicated in fixed task settings.
Shridhar et. al. (\textit{CLIPort}) \citep{CLIPort} proposed to combine the pretraining visual-language alignment capabilities of CLIP \citep{CLIP} with spatial precision of TransporterNets \citep{TransporterNR} to solve a range of language-conditioned manipulation tasks with efficient imitation learning. 
However, \textit{CLIPort} struggles to ground expressions that require reasoning about arbitrary visual concepts and complex relationships between objects.
Several other works propose disentangled pipelines for vision and action, with language primarily used to guide vision \citep{INGRESS, TellMeDave, holisticNlgrasp1, holisticNlgrasp, Blukis2020FewshotOG}.
The guiding process is implemented via relevancy clustering of LSTM-generated image-text features \citep{INGRESS} or element-wise fusion of images with sentence-wide text embeddings \citep{holisticNlgrasp1, holisticNlgrasp}.
Such holistic feature fusion approaches fall short to use richer object-word alignment, as motivated in the previous section.
Instead, in our work, we employ a neurosymbolic framework that utilizes explicit semantics about words and phrases and their correspondence to referring expressions in language commands.
In \citep{TellMeDave}, a parser is used to translate language instructions to formal programs operating on scene graphs, similar to our approach.
However, programs and scene representations are built with a constituency parser and heuristics respectively, thus being limited to the modeled vocabulary of concepts.
In our work, we use deep neural nets to do parsing and scene representation, as well as object-concept grounding, therefore entertaining benefits from both explicit semantics and representational strength of deep networks.

A plethora of recent works use large language models (LLMs) as semantic parsers to map natural language into Python-based programs composed of primitives~\citep{Instruct2ActMM,CodeAP,SocraticMC,VoxPoserC3, RobotGPTRM}, hence gaining open-vocabulary generalizabity due to the Internet-scale pretraining of the language model. However, such works rely heavily on prompt engineering and in-context examples to steer the LLM generation, making the system brittle and unreliable. 
Further, they require closed APIs or intense computational resources as part of the overall architecture, thus hindering its real-time applicability, which is essential in robotics.
Our works uses semantic parsing that is trained bottom-up from data, while maintaining open-vocabulary generalization by decoupling domain vocabulary from the DSL primitives.

\begin{figure*}[!t]
    \centering
    \includegraphics[width=1\textwidth]{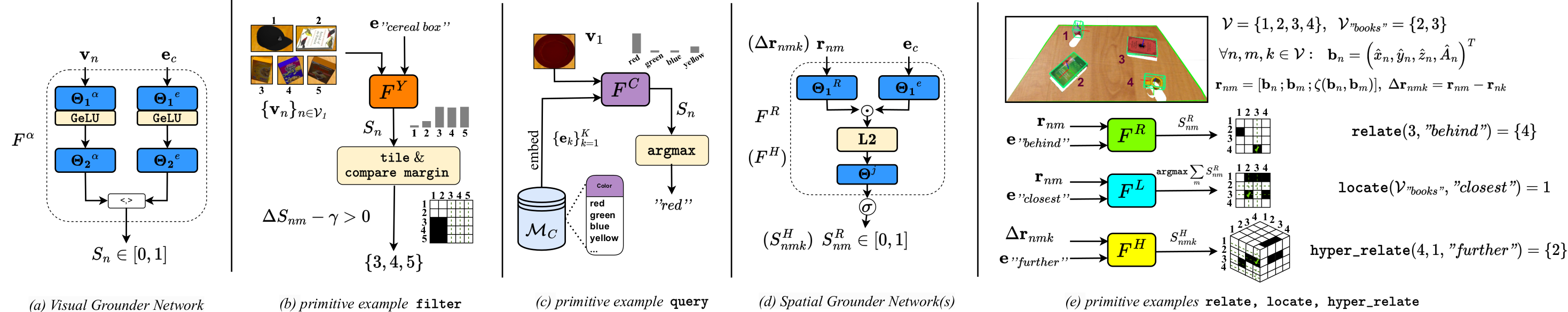}
    \caption{\footnotesize
    From left to right: (a) A Visual Grounder (VG) network is used to ground attribute concepts to object instances and vice versa. The program executor invokes VG to perform (b) \textit{filtering} and (c) \textit{querying} primitives by computing similarity scores for object-concept pairs. A Concept Memory $\mathcal{M_C}$ provides concept values and their embeddings to enable the VG to query over all encountered concept values. (d) A Spatial Grounder (SG) network is used to ground relation concepts to object pairs. The program executor invokes SG to resolve: (e) \textit{relations}, \textit{locations} and \textit{hyper-relations}. The relation and location primitives can be implemented via the relation grounder, while hyper-relations are resolved via a dedicated hyper-relation grounder network.}
    \vspace{-1mm}
    \label{fig:Fig3}
\end{figure*}

\section{Methodology}
Our architecture is comprised of four components: a) a scene encoder (\textit{hybrid}), b) a language parser (\textit{neural}), c) a dedicated language that implements a library of reasoning / action primitives, paired with a program executor (\textit{symbolic}) and d) a set of concept grounding modules (\textit{neural}).
Given a visual world state, the scene encoder constructs a scene graph representation that embeds object features as nodes and their spatial relations as edges. 
The language parser translates the input natural language query into the underlying program, expressed in our language, and the program executor executes it as a sequence of message passing steps in the extracted scene graph.
The concept grounders are used to interface words from the query that represent concepts with their matching objects in the scene representation.
The overall framework with a running example is illustrated in Fig.~\ref{fig:Fig2}. 

Since our focus in this work is the application of the system in open-vocabulary fashion, we make two important modifications to previous works.
First, we decompose the language parser into two sub-modules: a tagger network that replaces words in the query with their corresponding concept tags and a seq2seq network that translates the abstracted sequence to the final program.
This setup enables us to parse potentially new vocabulary, as long as the tagger has recognized the corresponding concept correctly.
Second, we replace hand-crafted relation primitives and attribute classification with object-concept grounding networks, opting to generalize to unseen concepts by leveraging the similarity semantics of pretrained word embeddings used to represent the concepts.

\subsection{Scene Encoder}
\label{scene_graph_rep}
Given an input RGB-D pair of images, we first apply an off-the-shelf object detector \citep{he2017mask} in RGB for instance segmentation and crop the $N$ detected object instances $\{I_n \in \mathbb{R}^{h_n \times w_n \times 3}\}_{n=1}^N$. 
Segmented objects are projected to 3D space using the camera intrinsics and approximated with a 3D bounding box $\mathbf{b}_n= \left ( {x_n} \; \,
{y_n}  \; \,
{z_n}  \; \,
{l^x_n} \; \,
{l^y_n}  \; \,
{l^z_n} \right )^T
 $, normalized according to the dimensions of the workspace.
The object boxes are used to mask object views from a top-down orthographic projection, providing a heightmap per object $\hat{I}_n \in \mathbb{R}^{h_n \times w_n}$.
We then construct a scene graph $\mathcal{G} = \left \{ \mathcal{V}, \mathcal{E}, \mathcal{X_V}, \mathcal{X_E} \right \}$ with nodes $\mathcal{V} = \left \{ 1, \dots, N \right \}$, edges $\mathcal{E} = \mathcal{V} \times \mathcal{V}$, node features $\mathcal{X_V} = \left \{  \mathbf{x}_{n}^{\mathcal{V}} = \left ( \mathbf{v}_n, \mathbf{g}_n  \right ), \; n \in \mathcal{V} \right \}$ and edge features $\mathcal{X_E} = \left \{ \mathbf{r}_{nm}, \; (n,m) \in \mathcal{E} \right \}$.

\textbf{Visual Encoder} We pass the cropped RGB images $I_n$ to a pretrained network $H:\mathbb{R}^{h_n \times w_n \times 3} \rightarrow \mathbb{R}^{D_v}$, comprised of up to the penultimate layer of an ImageNet \citep{imagenet} pretrained ResNet-50 \citep{resnet} and kept frozen.
The resulting feature maps are flattened to a single vector representation $\mathbf{v}_n = H(I_n^c)$ per object, of size $D_v$.

\textbf{Grasp Synthesis} We utilise a pretrained vision-based grasp synthesis network $G: \mathbb{R}^{h_n \times w_n } \rightarrow \mathbb{R}^5$ (e.g. \textit{GR-ConvNet} \citep{Antipodal}), that receives the input object heightmaps $\hat{I}_n$ and generates pixel-level masks $G(\hat{I}_n)=(\mathbf{\Phi},\mathbf{T},\mathbf{Q})_n \in \mathbb{R}^{3 \times h_n \times w_n}$, where $\mathbf{\Phi},\mathbf{W},\mathbf{Q}$ are each $\mathbb{R}^{h_n \times w_n}$ maps that contain the rotation with respect to the camera frame $\phi_n$, the grasp width $\omega_n$ and the grasp quality $q_n$ respectively.
We transform the grasp predictions in the world reference frame and select the center point $(u_n,v_n)^{\texttt{world}}$ that gives the grasp proposal with the best quality for each object $    \mathbf{g_n} := \max_{Q_n}\ G(\hat{I}_n)=\max_{Q_n} (\mathbf{\Phi},\mathbf{T},\mathbf{Q})_n$,
so that $\mathbf{g}_n = (u_n^{\texttt{world}}, v_n^{\texttt{world}}, \phi_n, \omega_n, q_n) \in \mathbb{R}^5$.

\textbf{Relation Encoder} We encode each pair-wise spatial relation between two objects $(n,m) \in \mathcal{E}$ with the concatenation of their normalized 3D boxes $\left [ \mathbf{b}_n \, ; \, \mathbf{b}_m \right ]$, as well as some binary relation features $\zeta(\mathbf{b}_n,\mathbf{b}_m) \in \{0,1\}$ that we extract from their boxes (e.g. $\left [ x_n + l_n^x/2 \leq   x_m - l_m^x/2 \right ]$, with $[\cdot]$ denoting evaluating the input condition for true/false).
Formally, each edge representation in our scene graph is given by:
\begin{equation*}
    \mathbf{r}_{nm}=\left [ \mathbf{b}_{n} \, ; \, \mathbf{b}_{m}  \,  ; \, \zeta(\mathbf{b}_n, \mathbf{b}_m) \right ]
\end{equation*}
We find that the extra binary features are essential for successfully grounding concepts such as \textit{``behind"}, as they contain more fine-grained relations about the object pair (e.g. overlap between objects in $x$-dimension). 
See Appendix~\ref{appendixSpatial} for more details.

\begin{table*}[!t]
    \centering
    \rowcolors{1}{}{lightgray}
     \resizebox{\textwidth}{!}{%
    \begin{tabular}{cccccc}
    \textbf{Reasoning} & \textbf{Primitive} &  \textbf{Concept Argument} ($\alpha$)  & \textbf{Type Signature} & \textbf{Semantics} & \textbf{Implementation} \\
    \toprule
     & \textbf{filter} &   Color, Material, Category & ($\mathcal{V}_{1}$: ObjSet, c: \texttt{str}) $\rightarrow$ ObjSet & \begin{tabular}{@{}c@{}}{Returns subset of objects with} \\ {given attribute concept value } \end{tabular} & $\{n \in \mathcal{V}_{1} \; | \;  \sum_{m \in \mathcal{V}_{1}}^{} [\gamma + F^{\alpha}(m, c) - F^{\alpha}(n, c) > 0] = 0 \}$  \vspace{1mm} \\
     \textit{visual} & \textbf{query} &  Color, Material, Category & $n_1$: Obj $\rightarrow$ \texttt{str}& \begin{tabular}{@{}c@{}}{Returns attribute concept} \\ {value for given object} \end{tabular}  & $\texttt{argmax}_c \; \{\sigma \left ( F^{\alpha} (n_1, c) \right ), \;  c \in \mathcal{M_C}[\alpha] \}$ \vspace{1mm}   \\ 
      & \textbf{same} &   Color, Material, Category & $n_1$: Obj $\rightarrow$ ObjSet & \begin{tabular}{@{}c@{}}{Returns subset of objects that have same} \\ {attribute concept value with given object} \end{tabular}  & $\texttt{filter} \left (\mathcal{V} - \{n_1\}, \; \; \texttt{query}(n_1)  \right )$ \vspace{1mm}   \\ 
    \hline
     & \textbf{relate} &  Relation & ($m_1$: Obj, $r$: \texttt{str}) $\rightarrow$ ObjSet & \begin{tabular}{@{}c@{}}{Returns subset of objects with given} \\ {relation value to given object} \end{tabular} & $\left \{ n \in \mathcal{V} \;  | \; \sigma(F^R(n, m_1, c)) \geq 0,5 \right \}$  \vspace{1.2mm} \\
     \textit{spatial} & \textbf{locate} &  Relation & ($\mathcal{V}_{1}$: ObjSet, $r$: \texttt{str}) $\rightarrow$ Obj & \begin{tabular}{@{}c@{}}{Returns object with most given relation} \\ {values from given object set} \end{tabular} & $\texttt{argmax}_n \; \left \{ F^L(n, c), \;  n \in \mathcal{V}_1 \right \}$  \vspace{1.2mm} \\
      & \textbf{hyper\_relate} &  Relation & ($m_1$: Obj, $m_2$: Obj, $r$: \texttt{str}) $\rightarrow$ ObjSet &  \begin{tabular}{@{}c@{}}{Returns subset of objects with given} \\ {relation value to given object pair} \end{tabular}  & $\left \{ n \in \mathcal{V} \;  | \; \sigma(F^H(n, m_1, m_2, c)) \geq 0.5 \right \}$  \vspace{1.2mm} \\
    \hline
      & \begin{tabular}{@{}c@{}}{\textbf{or},} \\ {\textbf{and}} \end{tabular}  &  - & ($\mathcal{V}_1$: ObjSet, $\mathcal{V}_2$: ObjSet) $\rightarrow$ ObjSet & \begin{tabular}{@{}c@{}}{Returns union/intersection of} \\ {two given object sets} \end{tabular}  & $\mathcal{V}_1 \cup   \mathcal{V}_2, \; \; \mathcal{V}_1 \cap \mathcal{V}_2$ \vspace{1.2mm} \\
      \begin{tabular}{@{}c@{}}{ }\\ {\textit{symbolic}} \end{tabular}  &  \begin{tabular}{@{}c@{}}{\textbf{exist},} \\ {\textbf{count}} \end{tabular}  &  - & $\mathcal{V}_1$: ObjSet $\rightarrow$ \texttt{bool/int} & Returns size of given object set & $\left [\left |\mathcal{V}_1  \right | > 0  \right ], \; \; \left |\mathcal{V}_1  \right |$ \vspace{1.2mm} \\
     &  \begin{tabular}{@{}c@{}}{\textbf{equal\_integer,}} \\ {\textbf{greater, less}} \end{tabular}  &  Integer & ($\nu_1$: \texttt{int}, $\nu_2$: \texttt{int}) $\rightarrow$ \texttt{bool} & Compares two given integers & $\left [\nu_1 = \nu_2  \right ], \;  \left [\nu_1 > \nu_2  \right ], \: \left [\nu_1 < \nu_2  \right ]$ \vspace{1.2mm} \\
     & \textbf{equal} &   Color, Material, Category  & ($c_1$: \texttt{str}, $c_2$: \texttt{str}) $\rightarrow$ \texttt{bool} &  Compares two given attribute concept values & $\left [c_1 = c_2  \right ]$   \\
     \toprule
    \end{tabular}%
  }
    \caption{\footnotesize
    The library of reasoning primitives included in our language. For brevity we don't enumerate all combinations of primitive and concept arguments, but illustrate the latter as a separate column. \textit{Visual} modules interface with visual grounders and the scene's visual features to reason about visual attributes. \textit{Spatial} primitives interface with spatial grounders to resolve spatial relations, absolute relations (locations) and hyper-relations. \textit{Symbolic} modules implement basic logic operations to incorporate integer and set semantics.}
    \label{tab:Tab1}
    \vspace{-4mm}
\end{table*}

\subsection{Language Parser}
The language parser consists of two sub-modules, a tagger network that identifies concepts in the input query and a seq2seq network for generating the program. 
To deal with potentially unseen vocabulary, the seq2seq network generates only the primitive functions of the overall program, whose arguments are restored from the query via a tag-conditioned  attention \textit{linear sum assignment} (LSA) module.

\textbf{Concept Tagger} We treat concept tagging similar to named entity recognition task in NLP \citep{ner}, where we map each word in the input query $\mathbf{w}_{1:T}$ to a tag $\mathbf{c}_{1:T}$, from a set of concept tags \{\textit{$\varnothing$, Category, Color, Material, Relation, Location, Hyper-Relation}\}.
Even though we can learn tagging with a shallow from-scratch network, we experimentally find that fine-tuning a pretrained language model achieves better generalization performance with less data (see Sec.~\ref{exp-gen}).
To that end, we fine-tune a pretrained \textit{distilBERT}$_{base}$ \citep{distilbert} model.
We use WordPiece tokenization \citep{wordpiece} and adopt the IOB scheme to deal with sub-word - tag misalignment (i.e., B - start of concept, I - continuation of concept, O - not a concept).
The tokens after the embedding layer $\mathbf{e}_{1:T}$ are cached, as they will be matched to arguments of the final program through the attention LSA module.
An example of tagging is given in Fig.~\ref{fig:Fig2} and more are shown in Appendix~\ref{appendixExamples}.

\textbf{Seq2Seq Encoder-Decoder} We replace words that are mapped to concepts with the corresponding tag and feed the replaced sequence as input to a RNN-based seq2seq network, enhanced with an attention layer between the encoder and decoder \citep{Bahdanau2014NeuralMT}. 
A two-layer Bi-GRU \citep{gru} of hidden size $D_{\pi}$ encodes the input sequence into hidden states $\mathbf{h}_t^{enc}=\texttt{Bi-GRU}(\mathbf{e_t}, \mathbf{h}_{t-1}^{enc})$ and a two-layer GRU decoder of hidden size $D_{\pi}$ generates the sequence of primitive functions $\mathbf{\pi}_{\tau}=\texttt{softmax}\left ( \mathbf{\Theta}_{\pi} \cdot \left [ \mathbf{h}_{\tau}^{dec} ; \mathbf{a}_{\tau} \right ] \right )$, selected through greedy decoding from the primitives library $\Pi$, using a linear layer $\mathbf{\Theta}_{\pi} \in \mathbb{R}^{D_{\pi} \times |\Pi|}$.
Here, $\mathbf{a}_t = \sum_{\tau}^{}\alpha_{t\tau} \mathbf{h}_{\tau}^{dec}, \; a_{t\tau} = \texttt{softmax}\left ( \mathbf{h}_t^{enc} \cdot \mathbf{\Theta}_{attn} \cdot \mathbf{h}_{\tau}^{dec} \right )$ denotes the weighted average of the attention scores over the hidden encoder states, where $\tau=1,...,\mathcal{T}$ the steps of the generated program.

\textbf{Tag-conditioned Attention LSA}
For each generated primitive function $\pi_{\tau}$ that receives concept arguments, only words tagged with the corresponding concept $C_{\tau}$ should be selected (e.g. $C_{\tau}$=\textit{Color} for $\pi_{\tau} = \texttt{filter\_color} $). 
We filter word tokens that satisfy this constraint and consider their normalized attention scores $\hat{a}_{t \tau} = \left \{ a_{t \tau} / \sum_{t}^{}a_{t \tau} \, |  \, c_t=C_{\tau} \right \}$.
Intuitively, the word $t$ whose hidden state was the most attended in order to generate the function $\pi_{\tau}$ corresponds to the argument of the function.
However, we experimentally find that when multiple instances of the same primitive appear in the program, not always the matching argument corresponds to the maximum attention score.
We then want to select the configuration of unique function-arguments pairs $(\tau,t)$ that maximizes the attention scores across functions $\sum_{\tau}{}\hat{a}_{t \tau}$, which is equivalent to the linear sum assignment problem, solved efficiently by the Hungarian matching algorithm \citep{Hungarian}.
The cached embedding $e_t$ is used as the argument for primitive $\pi_{\tau}$ for each selected pair.

\subsection{Concept Grounding}
\label{vlm}
The purpose of concept grounders is dual: (a) to match scene objects $n \in \mathcal{V}$ with attribute concepts (e.g. \textit{`bowl'} for category, \textit{`red'} for color, \textit{`plastic'} for material etc.) and vice-versa \textit{(visual)} using their visual features $\mathbf{v}_n$, and (b) to match object pairs $n,m \in \mathcal{E}$ with spatial concepts (binary relations, locations, and hyper-relations) based on their pair-wise relation features $\mathbf{r}_{nm}$ \textit{(spatial)}.
Fig.~\ref{fig:Fig3} illustrates the architecture of the grounder networks and how to run inference for implementing basic visual/spatial reasoning primitives of our library, namely \texttt{filter, query, relate, locate} and \texttt{hyper\_relate}.

\textbf{Visual Grounders (VG)} We implement a module $F^{\alpha}$ per attribute concept $\alpha \in$ \{Color, Material, Category\} that estimates a similarity score between a visual feature $\mathbf{v}_n$ of an object and a concept embedding $\mathbf{e_{c}}$, which corresponds to the (averaged) embedding(s) of a concept word(/phrase) $c$.
The similarity score is given by $F^{\alpha}(n,c) = < \mathbf{\hat{v}}_n, \mathbf{\hat{e}}_c >$, where:
\begin{equation*}
    \mathbf{\hat{v}}_n =  \frac{ \mathbf{\Theta_2^{\alpha}} \cdot \texttt{gelu}(\mathbf{\Theta_1^{\alpha}} \, \mathbf{v}_n)}{\left \|  \mathbf{\Theta_2^{\alpha}} \cdot \texttt{gelu}(\mathbf{\Theta_1^{\alpha}} \, \mathbf{v}_n) \right \|_2} , \;
    \mathbf{\hat{e}}_c = \frac{ \mathbf{\Theta_2^{e}} \cdot \texttt{gelu}(\mathbf{\Theta_1^{e}} \,\mathbf{e}_c) }{\left \|  \mathbf{\Theta_2^{e}} \cdot \texttt{gelu}(\mathbf{\Theta_1^{e}} \,\mathbf{e}_c)  \right \|_2}
\end{equation*}

\noindent with $\mathbf{\Theta_1^{\alpha}} \in \mathbb{R}^{D_j \times D_v}, \mathbf{\Theta_2^{\alpha}} \in \mathbb{R}^{D_j \times D_j}, \mathbf{\Theta_1^{e}} \in \mathbb{R}^{D_j \times D_e}, \mathbf{\Theta_2^{e}} \in \mathbb{R}^{D_j \times D_j}$ trainable matrices, $D_j$ the joint embedding dimension, $<,>$ the cosine similarity metric and $\texttt{gelu}$ the GeLU activation function \citep{GaussianEL}.

Following \citep{Mattnet}, we train VG using a hard margin hinge loss with in-batch sampling of negative object-concept pairs.
To do inference, we handle the two uses of VG separately. 
For \texttt{filter}, we need to select that subset of objects $n$ whose similarity difference from all other objects is not above a fixed margin $\gamma$, while for \texttt{query}, we want to select the concept value $c$ from the set of all possible attribute concepts (maintained in the concept memory module $\mathcal{M_C}$) that gives highest similarity with a single object $n_1$.
The exact formulas are given in Table~\ref{tab:Tab1}.

\textbf{Spatial Grounders (SG)} Resolving spatial relations comes in three flavours in our domain, namely: a) binary relations (e.g. \textit{``left of"}), that operate on pair-wise relation features $\mathbf{r}_{nm}$, b) absolute relations (i.e., \textit{locations} - e.g. \textit{``leftmost"}), that depend on the aggregation of all binary relations for a given object set $n \in \mathcal{V}_1$, and c) hyper-relations (e.g. \textit{``closer to/than"}), that operate on relative relation features $\Delta \mathbf{r}_{nmk} = \mathbf{r}_{nm} - \mathbf{r}_{nk}$ between a source $n$ and two target objects $m,k \in \mathcal{V}$.
As locations can be expressed via binary relations, we only need to implement two spatial grounding networks $F^R$ and $F^H$.
Formally:
\begin{equation*}
    F^{R}(n, m, c) = \mathbf{\Theta}_j^R \cdot \frac{\mathbf{\Theta}_1^R \cdot \mathbf{r}_{nm} \odot {\mathbf{\Theta}_1^e \cdot \mathbf{e}_{c}}}{\left \| \mathbf{\Theta}_1^R \cdot \mathbf{r}_{nm} \odot {\mathbf{\Theta}_1^e \cdot \mathbf{e}_{c}} \right \|_2}
\end{equation*}
\begin{equation*}
F^{L}(n, c) =  \sum_{m \in \mathcal{V}_1}^{} \sigma \left ( F^R(n, m, c) \right), \; n \in \mathcal{V}_1
\end{equation*}
\begin{equation*}
    F^{H}(n,m,k,c) = \mathbf{\Theta}_j^H \cdot \frac{\mathbf{\Theta}_1^H \cdot \Delta\mathbf{r}_{nmk} \odot {\mathbf{\Theta}_2^e \cdot \mathbf{e}_{c}}}{\left \| \mathbf{\Theta}_1^H \cdot \Delta\mathbf{r}_{nmk} \odot {\mathbf{\Theta}_2^e \cdot \mathbf{e}_{c}} \right \|_2}
\end{equation*}

\noindent where $\mathbf{\Theta_1^{R}} \in \mathbb{R}^{D_j \times D_R}, \mathbf{\Theta_1^{e}} \in \mathbb{R}^{D_j \times D_e}, \mathbf{\Theta_j^R} \in \mathbb{R}^{D_j \times 1}, \mathbf{\Theta_1^{H}} \in \mathbb{R}^{D_j \times D_H}, \mathbf{\Theta_2^{e}} \in \mathbb{R}^{D_j \times D_e}, \mathbf{\Theta_j^H} \in \mathbb{R}^{D_j \times 1}$ are trainable matrices, $D_j$ denotes the joint embedding dimension and $\odot$ the element-wise product. 
Spatial grounders are designed to produce binary matching scores between concepts and \textit{any} object pair of the scene, as in \citep{CMN}, hence the architectural difference between VG and SG networks. 
We train using a binary cross-entropy loss over all relations in all object pairs of each scene.

\subsection{Primitives and Program Execution}
\textbf{Primitives Library} We define our library of reasoning primitives $\Pi$ similar to the CLEVR domain \citep{clevr}, which we formally present in Table~\ref{tab:Tab1}.
The library includes two extra operational primitives, namely: a) \texttt{scene}, which initializes an execution trace returning all objects $\mathcal{V}$, and b) \texttt{unique}, which returns the object contained in a single-element object set.
Action primitives are terminal nodes in a program that control the robot arm via a custom first-party control API, whose implementation is orthogonal to our DSL.
In our implementation, the \texttt{grasp} primitive instructs the robot to grasp an input object $n$ using its grasp proposal $\mathbf{g}_n$ as the target end-effector pose for an inverse-kinematics solver.

\textbf{Program Executor} Primitives are developed as functions in a Python API. 
Our type system supports basic variable types, as well as two special types for representing an object and an object set through their unique indices in the scene graph nodes $\mathcal{V}$.
All functions share the same type system and input/output interface and thus can be arbitrarily composed in any order and length.
As in \citep{ns-vqa}, branching structures due to double argument primitives (e.g. \texttt{and}) are handled via the usage of a stack, allowing program execution as a chain of module calls, each receiving as input the output of the previous step and accessing the stack in case of double arguments.
Whenever there is a type mismatch between expected and retrieved inputs/outputs, a suitable response is returned, enforcing interpretability by explaining to the user which reasoning step failed. 
To speed up computation, we first group all program steps that require concept grounding to do a single batched forward pass per grounder, and mask the network predictions during execution according to the previous steps.

\subsection{Training Paradigm}
\label{training}
The training process entails two optimization objectives: a) the correctness of the parsed program and b) object-concept matching of the concept grounders. 
Following insights from prior works \citep{ns-cl}, we train using a curriculum learning approach. 
In particular, we first train the grounder modules to ground attribute concepts to objects \textit{(VG)} and spatial concepts to object pairs (\textit{SG}).
To that end, we isolate input/output pairs from filtering, querying and relation-based operations from the execution traces of our dataset's program annotations and express them as binary masks over the graphs nodes (VG) / edges (SG). 
We train the grounders on the checkpoint datasets and freeze their weights for the following steps.
For language parsing, we first train the concept tagger on a small split of tagged queries and then the entire language parser objective following \citep{ns-vqa}.
First, we select a small diverse split of the training data, sampling uniformly from all different templates, and train using the ground truth programs with a cross-entropy loss.
Finally, we combine the language parser with the grounders and the program executor and train the system end-to-end in the remaining scenes with REINFORCE \citep{REINFORCE}, using only the correctness of the executed program as the reward signal.

\section{Experiments}
We structure our experimental evaluation as follows: First (Sec.~\ref{exp-data}), we present the details of the synthetic dataset generation and the collected real-world dataset.
In Sec.~\ref{exp-sim} and ~\ref{exp-gen}, we evaluate the visual reasoning capabilities of the proposed model through VQA, where we compare our approach with previous baselines in terms of accuracy, sample-efficiency and generalization to unseen vocabulary.
In Sec.~\ref{exp-real}, we study the transfer performance of our method in real scenes via few-shot fine-tuning of our visual grounder network.
In Sec.~\ref{exp-grasp}, we integrate our method with a robot framework and perform end-to-end experiments for an interactive object grasping scenario, where we examine the distribution of failure modes across system components in scene-instruction pairs with increasing complexity.
Finally (Sec.~\ref{exp-grasp}), we show that our method can be extended to more manipulation tasks via few-shot fine-tuning of the language parser.
\subsection{Datasets}
\label{exp-data}
We present the synthetic and real versions of the dataset we release, termed: \textit{\textbf{H}ousehold \textbf{O}bjects placed in \textbf{T}abletop \textbf{S}cenarios} \textit{(HOTS)}.
We refer the reader to Appendix~\ref{appendixData} for more details on both versions.

\textbf{SynHOTS} We collect from available resources a catalogue of $58$ 3D object models from five types (fruits ($6$), electronics ($4$), kitchenware ($18$), stationery ($17$) and edible products ($13$)), organized into $25$ object categories, $10$ color and $8$ material concepts. 
As we strive for natural interaction, we also include instance-level object annotations according to their brand, variety or flavour (e.g. \textit{``Coca-Cola``} vs \textit{``Pepsi``}, \textit{``strawberry juice``} vs. \textit{``mango juice``} etc.)
We render synthetic scenes in the Gazebo environment \citep{Gazebo} and generate around $8k$ training and $1.6k$ validation RGB-D pairs, additionally equipped with parsed semantic scene graphs, containing all location, grasp, attribute and relation information for each object.
For annotating our scene graphs with language data, we develop on top of the CLEVR generation engine \citep{clevr} and produce language-program-answer triplets from synthetic task templates by sampling concepts from the scene graphs.
We extend the standard VQA templates of CLEVR to incorporate our designed DSL, as well as extra REF and grasping tasks, ending up with $11$ distinct task families, spawning a total of $295$ task templates, with rich variation in phrasing / syntax.
For the VQA task (\textit{SynHOTS-VQA}), we instantiate $66$ templates for each scene (6 per task family) and generate around $500$k training and $100$k validation question-program-answer samples.

\begin{figure}[!t]
\subfloat{\includegraphics[width=0.48\textwidth]{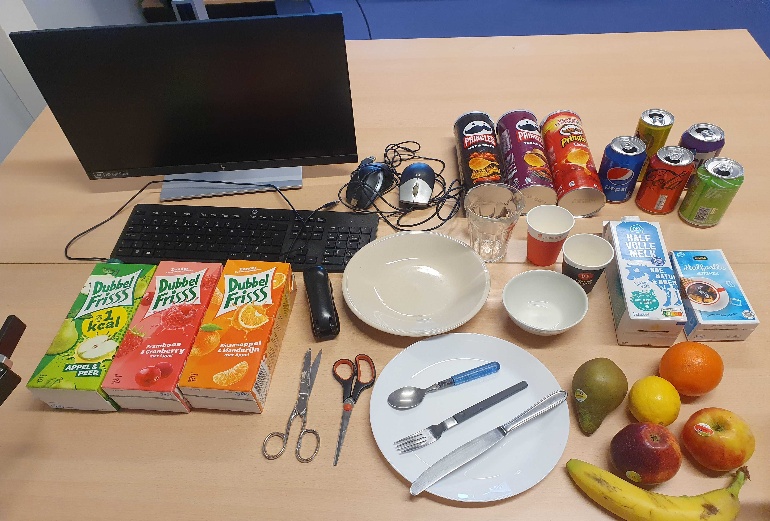}}
\hspace{0.01\textwidth}
\subfloat{\includegraphics[width=0.48\textwidth]{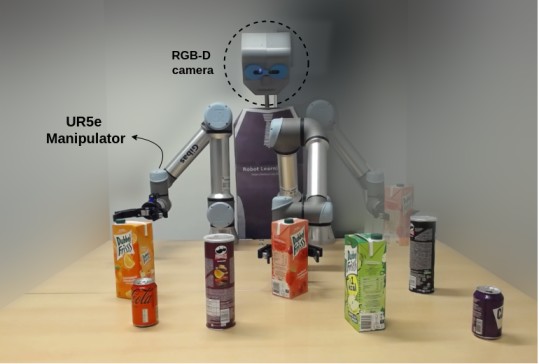}}
\caption{\footnotesize
A subset of the object catalogue included in the HOTS dataset (left) and an image of our real robot setup from the opposite perspective (right).}
\label{fig:Fig4}
\end{figure}

\textbf{HOTS} In order to evaluate the performance of our model in natural scenes, we record a dataset of real RGB-D images captured from a robot's camera. 
The real household objects used in this dataset, together with our dual-arm robot setup, are shown in Fig.~\ref{fig:Fig4}. The object catalogue is a subset of the synthetic one but includes a few novel attributes, for a total of $48$ object instances with $25$ category, $10$ color, and $7$ material concepts in $108$ unique scene configurations.
$22$ scenes that provide a fair representation of all concepts are held out for potential fine-tuning experiments, and the $86$ remaining scenes are used for testing.
We extract scene graphs and repeat the language-program-answer data generation step as in simulation, ending up with $5\,676$ scene-question pairs.

\subsection{VQA Evaluation in Simulation}
\label{exp-sim}
\textbf{Setup} We compare our method with three holistic (\citep{san, RN, FILM}) and the original NS-VQA \citep{ns-vqa} baseline.
The holistic models are trained using the implementation and hyper-parameters from \citep{FILM} and NS-VQA is a replica of the original work, with the executor component adapted to incorporate our primitives library.
We use a ResNet50 \citep{resnet} backbone for visual feature extraction and sample $4\,000$ images from our dataset to train the NS-VQA attribute classifiers and our grounders.
NS-VQA and our method are pretrained with $300$ programs sampled uniformly from all question families and fine-tuned with REINFORCE for the rest of the dataset. 
We note that our method additionally pretrains the tagger component of our parser with $500$ question-tag pairs.
We use Adam optimizer with batch size of $64$ and train for $2k$ iterations in pretraining and $2M$ iterations in REINFORCE stage, using learning rates of $3 \cdot 10^{-4}$ and $10^{-5}$ respectively. The reward is maximized over a constant baseline with a decay weight of $0.9$.

\textbf{Accuracy} We report results in \textit{SynHOTS-VQA} validation split in Table~\ref{tab:Tab2}, organized by question type. 
The metric used is final VQA accuracy, measured as top-1 prediction in the case of holistic and the correctness of the executed program in the case of neurosymbolic baselines.
Our model achieves near-perfect accuracy and is consistently above all holistic baselines across all question types, with the most significant margin in counting questions. 
Compared to NS-VQA, our approach achieves on-par performance, with a small drop due to the reformulation of the primitives library to be vocabulary-agnostic and the addition of the concept tagging bottleneck. 
We show in the next section that this drop is a favorable trade-off between performance in validation (seen) and generalization-test (unseen) splits.

\begin{figure*}[!t]
\begin{tabular}{c}
    \includegraphics[width=1\textwidth]{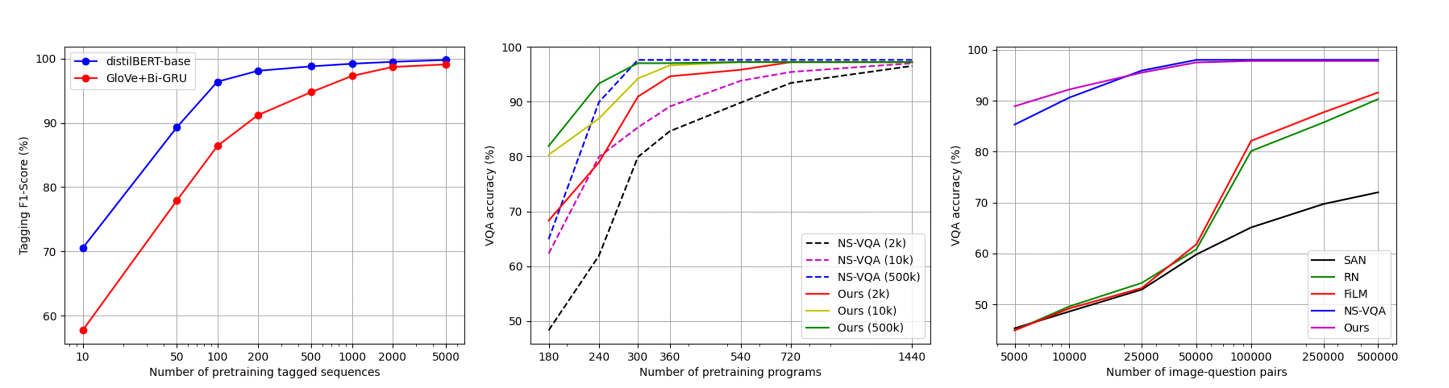}
\end{tabular}
\caption{\footnotesize
Sample-efficiency experiments on \textit{SynHOTS-VQA}. From left to right: \textit{(left):} F1-score of concept taggers vs. number of tagged annotations used during pretraining, \textit{(middle):} VQA accuracy vs. number of pretraining programs; different curves indicate different amounts of data used at the REINFORCE stage, \textit{(right):} VQA accuracy vs. number of training question-answer pairs; NS-VQA and our method are pretrained with $500$ programs.}%
\label{fig:Fig5}
\end{figure*}

\textbf{Sample-efficiency} We further analyze the sample-efficiency of our method compared to baselines in Fig.~\ref{fig:Fig5}, both in terms of pretraining and REINFORCE fine-tuning.
Regarding tagger pretraining, we see that with a powerful pretrained model such as \textit{distilBERT} \citep{distilbert} we achieve $99.8\%$ F1-score on the validation tags with only $500$ samples.
A GRU baseline with pretrained GloVe embeddings \citep{GloVe} needs $2k$ samples to achieve the same performance.
Regarding supervised pretraining, we see similar performance between NS-VQA and our method, with the latter being more efficient in weaker REINFORCE supervision ($2k$ and $10k$ question-answer pairs). 
We believe this result is due to our two-step parser implementation, as e.g. for as little as $180$ programs, the training examples most likely do not sufficiently cover the concept vocabulary of the domain for the NS-VQA parser, whereas in our method concept words are replaced by tags, which suffice in number.
Finally, our method is the most sample-efficient in terms of required question-answer pairs, with a significant gap compared to holistic approaches, which comes at the cost of just a few hundred question-programs annotations for supervised pretraining. 

\begin{table}[!t]
     \vspace{-2.5mm}
    \centering
    \resizebox{\textwidth}{!}{%

    \begin{tabular}{ccccccccc}
    \toprule
    \textbf{Method} & \textbf{Count} & \textbf{Exist} & \begin{tabular}{@{}c@{}}\textbf{Compare} \\ \textbf{Number}\end{tabular} & \begin{tabular}{@{}c@{}}\textbf{Compare} \\ \textbf{Attribute}\end{tabular} & \textbf{Query} & \textbf{REF} & \textbf{Overall} \\
    \midrule
     CNN-LSTM-SAN \citep{san} & 58.9 & $77.1$ & $73.9$ & $70.2$ & $79.8$ & - & $72.0$ \\
     CNN-LSTM-RN \citep{RN} & $86.3$ & $93.7$ & $87.05$ & $91.6$ & $92.8$ & - &  $90.3$ \\
     CNN-GRU-FiLM \citep{FILM}& $88.3$ & $93.4$ & $89.35$ & $92.9$ & $93.2$ & - &  $91.4$ \\
     \midrule 
     NS-VQA \citep{ns-vqa} & $98.6$ & $99.4$  & $98.1$ & $99.6$ & $95.6$ & $99.0$ & $98.2$ \\
     Ours  & $95.5$ & $97.9$ & $97.0$ & $99.7$ & $94.0$ & $99.6$ & $96.9$ \\
     \bottomrule
    \end{tabular}%
  }
    \caption{\footnotesize
    VQA accuracy (\%) per question type and overall for the validation split of our synthetic dataset. The \textit{REF} column denotes referring expression questions, that do not apply to baselines that are trained for closed-VQA. }
    \label{tab:Tab2}
    \vspace{-1mm}
\end{table}

\begin{table}[!b]
     \vspace{-2.5mm}
    \centering
     \resizebox{\linewidth}{!}{%

    \begin{tabular}{cc@{\hskip 0.3in}ccc@{\hskip 0.3in}cc}
    \toprule
    \multicolumn{2}{c}{\textbf{Setup}} &
      
    \multicolumn{3}{c}{\textbf{HOTS-Perc.}} &
    \multicolumn{2}{c}{\textbf{HOTS-Reas.}} \\
     Method & \#Data & Cat & Col & Mat &  REF & VQA \\
     \midrule
      GT & - &  $100.0$ & $100.0$ & $100.0$ & $96.8$ & $96.1$  \\
      VG-\footnotesize{no-pretrain}& \textit{full}& $92.9$ & $92.1$ & $90.4$ & $90.1$ & $88.2$ \\
     \midrule
     VG-\footnotesize{pretrain} & $0$  & $34.7$ & $40.4$ & $13.9$ & $26.6$ & $29.1$ \\
     VG-\footnotesize{pretrain} & $1$  & $43.2$ & $44.4$ & $60.8$ & $45.4$ & $47.7$ \\
     VG-\footnotesize{pretrain} & $5$ & $62.5$ & $67.6$ & $73.1$ & $66.1$ & $61.9$ \\
     VG-\footnotesize{pretrain} & $20$  & $90.5$ & $89.9$ & $94.4$ & $89.8$ & $86.6$ \\
     VG-\footnotesize{pretrain} & \textit{full} & $93.4$ & $91.8$ & $95.7$ & $90.9$ & $88.1$ \\
     \bottomrule
  \end{tabular}%
  }
  \caption{\footnotesize
  Top-1 accuracy (\%) for classifying attributes - \textit{category} (Cat), \textit{color} (Col) and \textit{material} (Mat) - as well as execution accuracy for end-to-end REF and VQA tasks in annotated scenes of our HOTS dataset. \textit{GT} denotes using ground truth attribute labels from scene graphs. The \#Data column denotes the number of fine-tuning examples per object instance for the VG.}
  \label{tab:Tab3}
\end{table}

\begin{table}[!t]
     \vspace{-2.5mm}
    \centering
    \resizebox{\textwidth}{!}{%

    \begin{tabular}{lccccc}
    \toprule
    \multirow{2}{3em}{\textbf{Method}}  & \textbf{Unseen} & \textbf{Unseen} & \textbf{Unseen}  & \multirow{2}{3em}{\textbf{Open}} & \multirow{2}{3em}{\textbf{Overall}} \\
      & \textbf{Category} & \textbf{Color} & \textbf{Material} &  &  \\
    \midrule
    NS-VQA (\footnotesize{lang$\rightarrow$prog})  & $30.4$  &  $13.1$ & $22.7$  & $29.9$   &  $28.8$ \\
    \multicolumn{1}{r}{\footnotesize{\textit{w/ GT-Perc.}}}  & $38.6$  &  $19.0$ & $29.6$  & $36.3$   &  $35.2$ \\
     \midrule 
     Ours (\footnotesize{lang$\rightarrow$tag$\rightarrow$prog})  & $68.4$ & $58.2$   & $78.4$   & $86.6$   &   $77.1$   \\
     \multicolumn{1}{r}{\footnotesize{\textit{w/ GT-Perc.}}}  & $73.2$  & $64.0$  &  $83.0$ & $93.0$  & $87.6$   \\
     \multicolumn{1}{r}{\footnotesize{\textit{w/ GT-Perc. + GT-Tags}}}   &  $94.1$ & $82.9$ & $95.1$ & $95.1$  & $94.8$    \\
     \bottomrule
    \end{tabular}%
  }
    \caption{\footnotesize
    VQA accuracy (\%) in generalization-test splits that contain questions with unseen vocabulary describing \textit{Category}, \textit{Color} and \textit{Material} concepts. \textit{Open} denotes the use of an unseen word to describe an object at instance-level. We note that a question might contain unseen words from multiple categories, so the \textit{Overall} column does not correspond to the average.}
    \label{tab:Tab4}
    \vspace{-2mm}
\end{table}

\subsection{Generelization to Unseen Vocabulary}
\label{exp-gen}
In this subsection we wish to evaluate the generalization performance of our model in unseen vocabulary, i.e. testing in words to describe concepts that were not part of the training data. 
We conduct experiments in four splits, three for unseen attribute concepts and one \textit{Open}, where we use unseen instance-level descriptions of a unique object in the scene (e.g. \textit{``Coca-Cola". ``mango juice"} etc. - check Appendix~\ref{appendixData} for full list). 
We perform several ablation experiments where we either use attribute labels from ground truth scene graphs or the actual perception pipeline (classifiers for NS-VQA and VG for our model), as well as ground truth tags instead of taggers predictions. 
The purpose here is to decompose the error rate to tagger, seq2seq and VG errors, in order to understand which module is the main bottleneck for generalization.
For a fair comparison with the NS-VQA baseline, for this experiment we initialize and freeze the word embedding layers of both methods with GloVe \citep{GloVe} and use our from-scratch GRU tagger baseline (pretrained with $2k$ question-tag pairs).
Results are summarized in Table~\ref{tab:Tab4}.
The vocabulary-aware baseline of NS-VQA fails to parse unseen concept words, as they are not part of the training data, while our approach achieves significantly higher accuracy, with near-perfect results when evaluating only the seq2seq network with ground truth perception and tags.
We identify VG as the main generalization bottleneck ($17.7\%$ overall accuracy drop when adding VG vs. $7.2\%$ when adding the tagger), with still however a large margin from NS-VQA.

\subsection{Adapting to Real Scenes}
\label{exp-real}
In this subsection, we wish to assess the transferability of our model in natural scenes by evaluating visual reasoning performance in the HOTS dataset.
We highlight that unlike holistic approaches, which require both vision and text data to be adapted, the modular nature of our approach allows us to bridge the sim-to-real gap solely in the vision domain, only fine-tuning the VG in real images and transferring the language parser \textbf{without} any further training.
We evaluate in two setups, namely: a) \textit{HOTS-Recognition}, where we only test the visual pipeline by treating attribute noun phrases as class labels like in the classification task, and b) \textit{HOTS-Reasoning}, where we test the end-to-end system for REF and VQA tasks separately.
For the first split, we use VG for querying attribute concepts of input object images and report the percentage of correct top-1 predictions as accuracy.
For this experiment, we directly provide the concept embeddings of all possible attribute tags from our concept memory.
We initialize the VG with the synthetic pretraining weights and fine-tune in different amounts of training examples per object instance (\textit{1, 5, 20}), as well as in \textit{full} dataset. 
A no-pretrained VG baseline that is only trained in real data is also included.
Results are summarized in Table~\ref{tab:Tab3}.
We observe that our method can be efficiently transferred to real scenes, as $20$ labeled examples per object instance achieves very similar performance to training from scratch in the entire dataset, both in the attribute recognition as well as in the end-to-end reasoning tasks.
We identify that the main bottleneck here is not the sim-to-real gap but the inclusion of unseen attribute concepts in \textit{HOTS} compared to \textit{SynHOTS}, which require more data as they are effectively learned from scratch by the visual grounder.

\begin{figure}[!t]
    \centering
    \includegraphics[width=1\textwidth,height=10cm,keepaspectratio]{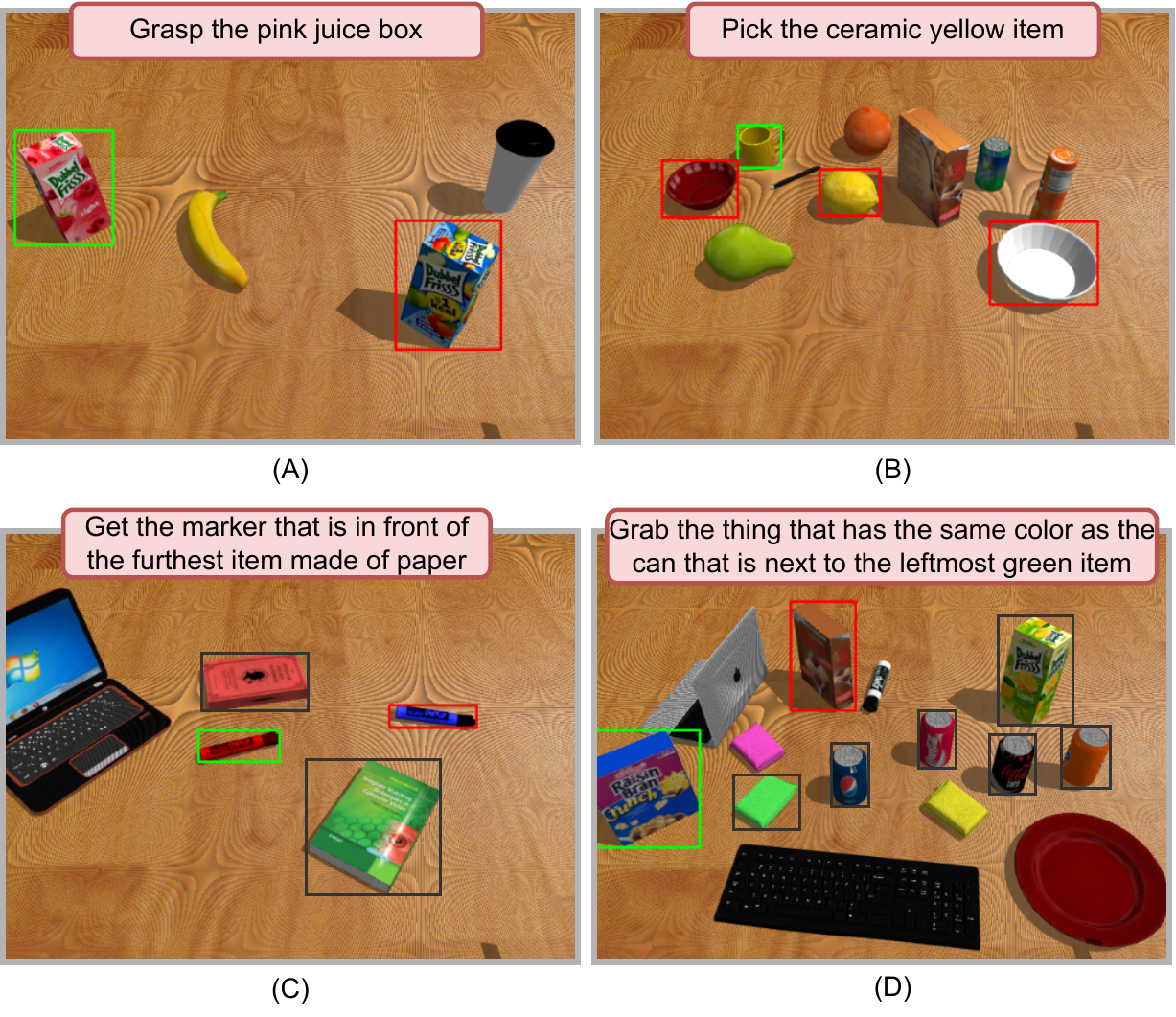}
\caption{ \footnotesize  
Example trials from the four splits used for simulated grasping experiments, namely:  (A) Scattered scenes - simple queries, (B) crowded scenes - simple queries, (C) scattered scenes - complicated queries, and (D) crowded scenes - complicated queries. The green box denotes the target item, red denotes a distractor item of the same attribute and the dark box denotes all items involved in the reasoning process.}
         \label{fig:Fig6}
         \vspace{-4mm}
\end{figure}

\subsection{Interpretable Interactive Object Grasping}
\label{exp-grasp}

\begin{figure*}[t]
    \centering
    \includegraphics[width=1\textwidth]{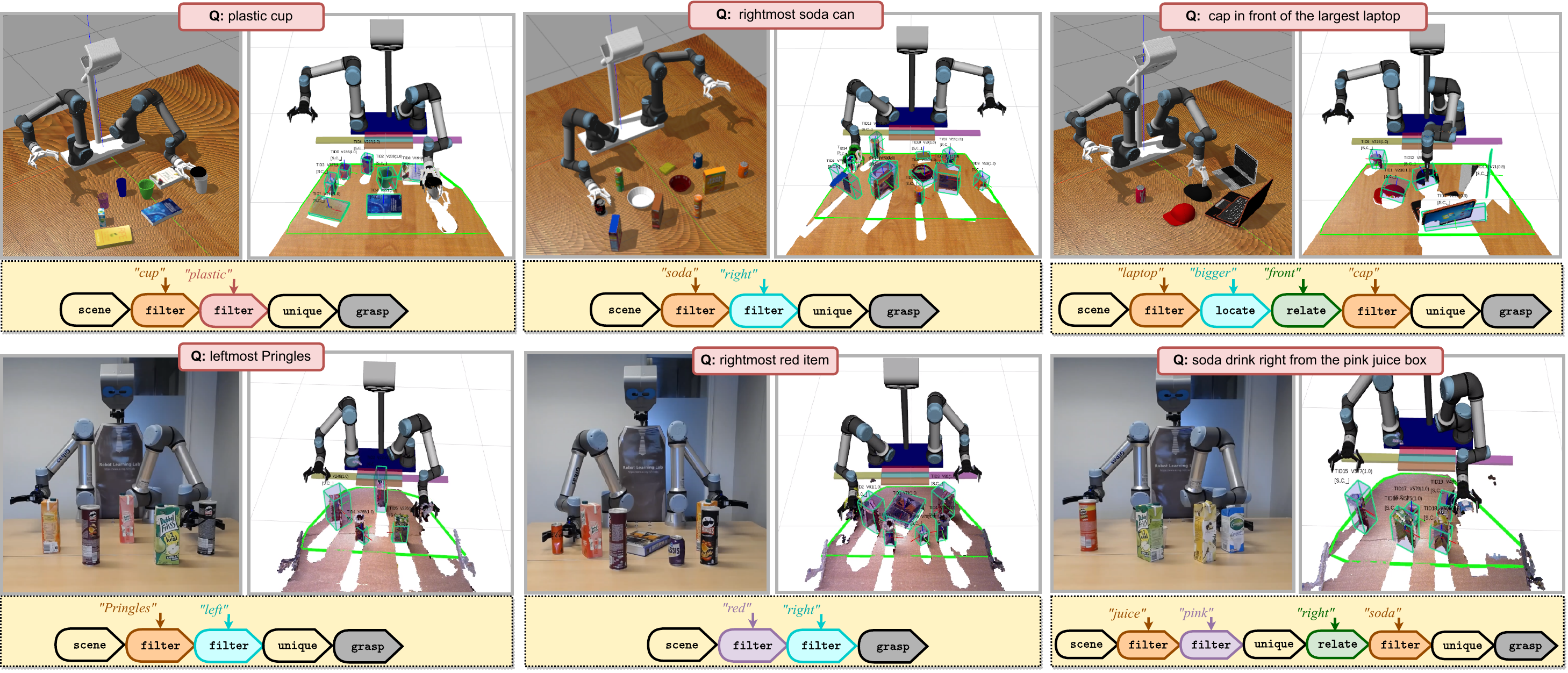}
    \caption{\footnotesize
    A sequence of snapshots capturing the setup of our robot framework in Gazebo (\textit{top)} and in a real-world environment (\textit{bottom)}. We generate a random scene and command the robot to grasp a specific item with a text instruction, referring to attributes / relations between objects \textit{(in pink)}. In the snapshots, we demonstrate the robot during the picking action (\textit{each-left)} and the localization results in RViz (\textit{each-right}), as well as the parsed program corresponding to the query \textit{(each-bottom)}.}
    \label{fig:Fig7}
    \vspace{-3mm}
\end{figure*}

In this subsection, we integrate our method with the grasping pipeline of \citep{Oliveira20163DOP} and evaluate its end-to-end behavior for an interactive object grasping task.
An illustration of the setup and experiments is given in Fig.~\ref{fig:Fig7}.
We conduct several trials, in which we randomly place objects on a table and instruct the robot to grasp an object in real time. 
The scenes always include distractor objects of a same attribute, requiring the user to use other attributes and/or spatial relations to uniquely refer to the goal object.
We note that the instructor is not limited to the concept vocabulary of our domain and can use arbitrary phrasing, potentially outside the syntax of our scripted templates.
The interpretable nature of our system allows us to examine the parsed program execution traces and diagnose the source of failures, including: a) \textit{perception}, where there is either a localization error or a grounder has given an incorrect match, b) \textit{reasoning}, where the parsed program is incorrect, or c) \textit{grasping}, where the grasping fails (e.g. due to collisions).

\begin{table}[t]
\vspace{-2.5mm}
    \centering

         \resizebox{\linewidth}{!}{%
    \begin{tabular}{cccccccc}
    \toprule
    \multicolumn{3}{c}{\textbf{Split}}  & \multirow{2}{3em}{\textbf{\#Trials}} & \multirow{2}{2.5em}{\textbf{\#Fail.}} & \multirow{2}{4em}{\textbf{\#Perc.Fail.}} & \multirow{2}{4em}{\textbf{\#Reas.Fail.}} & \multirow{2}{4em}{\textbf{\#Gr.Fail.}} \\
     \textit{env} & \textit{query} & \textit{scene} & & & & & \\
    \midrule
    & simple & scattered & 50  & 4 (8.0\%)  & 2 (4.0\%) & 0 (0.0\%)& 1 (2.0\%) \\ 
     & simple & crowded & 50 & 8 (16.0\%) & 4 (8.0\%) & 0 (0.0\%) & 4 (8.0\%) \\
    \textbf{Sim} & complex & scattered & 50 & 8 (16.0\%) & 4 (8.0\%) & 2 (4.0\%) & 2 (4.0\%) \\
     & complex & crowded & 50 & 19 (38.0\%) & 10 (20.0\%) & 3 (6.0\%) & 6 (12.0\%) \\
     \multicolumn{3}{c}{\textit{\hspace{1cm} total}}  & 200 & 39 (19.5\%) & 20 (10.0\%) & 5 (2.5\%) & 13 (6.5\%) \\
        \midrule
     & simple & scattered & 3 & 0 (0.0\%) & 0 (0.0\%) & 0 (0.0\%) & 0 (0.0\%) \\
     & simple & crowded & 3 & 1 (33.3\%)  & 1 (33.3\%) & 0 (0.0\%)& 0 (0.0\%) \\
    \textbf{Real} & complex & scattered & 3 & 0 (0.0\%) & 0 (0.0\%) & 0 (0.0\%) & 0 (0.0\%) \\
     & complex & crowded & 3 & 2 (66.6\%) & 1 (33.3\%) & 1 (33.3\%) & 0 (0.0\%) \\
     \multicolumn{3}{c}{\textit{\hspace{1cm} total}}  & 12 & 3 (25.0\%) & 2 (16.6\%) & 1 (8.3\%) & 0 (0.0\%) \\
     \bottomrule
    \end{tabular}%
    }
    \caption{\footnotesize
    Evaluating the system for an interactive object grasping task in synthetic \textit{(top)} and real \textit{(bottom)} scenes of incremental query and scene complexity. The interpretable nature of our approach allows us to decompose the failure modes across the different modules.}
    \label{tab:Tab5}
      \vspace{-1mm}
\end{table}

We report results in synthetic scenes separated in four splits, comprised of different levels of scene and query complexities (see Fig.~\ref{fig:Fig6}).
We generate 10 scenes per split and conduct 5 trials for each, for a total of $200$ scene-instruction pairs.
For the real experiments, we conduct a total of 12 trials using objects from the HOTS dataset and the adapted visual pipeline of the previous section. 
Results are summarized in Table~\ref{tab:Tab5}. 
We observe that in both setups the averaged error rate is similar ($20-25\%$), with the reasoning module being the most robust to grasping instructions across all trials.
Exceptions are a few queries in cases of complex question splits.
Such failures are mostly due to unique phrasing of the instruction by the human instructor, with one case of referring to an unknown spatial concept (e.g. \textit{``between"}).
Perception errors occur more frequently in the crowded scene setup, due to partial views of objects leading to occlusion.
We include a video with robot demonstrations as supplementary material.
The overall results showcase that the system can indeed serve as an accurate and interpretable interactive robotic grasper, while having relative robustness to free-form instructions.

\subsection{Extending to More Manipulation Tasks}
\label{exp-tasks}
\begin{figure}[t]
    \centering
    \includegraphics[width=1\textwidth]{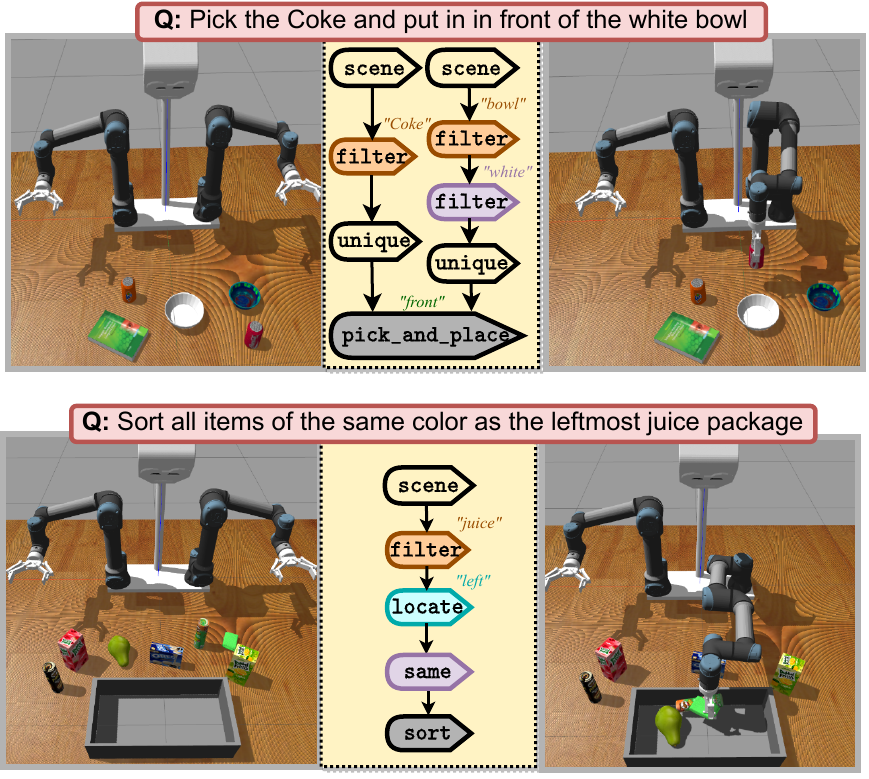}
    \caption{\footnotesize
    Extending to more manipulation tasks: \textit{(top):} pick an object and place it relative to another object, and \textit{(bottom):} sort all objects in a pre-defined container according to a reference object. }
    \label{fig:Fig8}
\end{figure}

In this subsection, we explore how efficiently our model can adapt to more complex manipulation tasks beyond grasping.
To that end, we implement two extra control primitives, which like \texttt{grasp}, act as terminal nodes in the parsed program, receiving unique indices of objects to manipulate and control the arm based on the grasp poses of the objects with an IK-solver.
In particular, we implement: a) \texttt{pick\_and\_place}, which receives two object inputs and a relation concept argument that map to what to pick, where to place, and how to place it respectively, and b) \texttt{sort}, which receives a set of objects to sort into a fixed container item (see Fig.~\ref{fig:Fig8}).
We structure new templates for these tasks and generate $10$ instruction-program pairs for $50$ novel synthetic scenes with the same constraints as the grasping task, for a total of $500$ pairs.
We fine-tune our language parser in the new instructions (while keeping the rest of the system fixed) and report results in Table~\ref{tab:Tab6}, using the same setup as the previous section for 100 trials per task in simulation.
As with grasping, we observe that the reasoning module is robust in query complexity and task success is limited only by perception and grasping modules, in cases of crowded scenes.
We further integrate policies obtained through behavioral cloning as control primitives and demonstrate more complex, long-horizon manipulation tasks in our supplementary material.

\begin{table}[t]
\vspace{-2.5mm}
    \centering

         \resizebox{\linewidth}{!}{%
    \begin{tabular}{ccccccc}
    \toprule
         \multicolumn{2}{c}{\textbf{Split}}  & \multirow{2}{3em}{\textbf{\#Trials}} & \multicolumn{2}{c}{\textbf{pick and place}} & \multicolumn{2}{c}{\textbf{sort by reference}} \\
      \textit{query} & \textit{scene} & & \textit{Pars.Acc} & \textit{Succ.Rate} & \textit{Pars.Acc} & \textit{Succ.Rate} \\
    \midrule
     simple & scattered & 25  & $100.0$ & $92.0$ & $100.0$ & $96.0$ \\ 
      simple & crowded & 25 & $100.0$ & $80.0$ & $100.0$ & $76.0$ \\
    complex & scattered & 25 & $96.0$ & $88.0$ & $88.0$ & $92.0$ \\
      complex & crowded & 25 & $96.0$ & $80.0$ & $96.0$ & $64.0$ \\
       \multicolumn{2}{c}{\textit{total}} & 100 & $98.0$ & $85.0$ & $96.0$ & $82.0$ \\
        \bottomrule
    \end{tabular}%
    }
    \caption{\footnotesize
    Evaluating the system for interactive \texttt{pick\_and\_place} and \texttt{sort\_by\_reference} manipulation tasks in synthetic scenes of incremental query and scene complexity. Results include parsing accuracy (\%), measured as the percentage of correctly generated programs for the input query, as well as success rate (\%) of the overall behavior, incl. perception and grasping modules.}
    \label{tab:Tab6}
      \vspace{-1mm}
\end{table}

\subsection{Comparisons with Foundation Models}
\label{exp-tasks}
In this section we explore the comparative performance, as well as integration potential, of our neurosymbolic framework with approaches relying on modern foundation models \citep{TowardsOG, OCIDVLG}, such as LLMs \citep{GPT4} and VLMs \citep{CLIP} for zero-shot semantic parsing and grounding.
To that end, we perform three experiments studying relative performance in different parts of the pipeline, including parsing, grounding and end-to-end grasping from a language instruction. 
To strengthen our evaluation we also utilize the dataset OCID-VLG \citep{OCIDVLG}, which provides referring expression queries, accompanied with parsed program and ground-truth mask and grasp annotations for 1763 unique scenes from the OCID dataset \citep{OCID}. 
To examine out-of-distribution generalization, we use the \textit{novel-classes} split provided from the authors, which includes object category objects unseen during training. We note that NS-MAN is not equipped to deal with novel concept types, since it is trained only on those existing in the training data. However, we conduct evaluations in this split to explore trade-offs between foundation model-based and our neurosymbolic approach, as well as potential for integrating the two.

\noindent \textbf{Semantic parsing} We compare our work with an LLM parser based on \texttt{gpt-4o} \citep{GPT4}, prompted with 8-shot examples from question-program pairs of our dataset, both for the REF task of SynHOTS and for OCID-VLG. We report final accuracy after program execution, using ground-truth perception in all methods, in order to understand the upper bound performance with perfect perception. In the case of OCID-VLG, we train from scratch our framework similar to SynHOTS.
Results are in Tab. \ref{tab:Tab7}.
We observe that with 8 in-context examples we can guide the LLM to provide perfect parsing in all splits, with errors only in identifying object instance names unseen in the prompt in OCID-VLG test (e.g. \textit{`Feh package'} as tissue box category).
Naturally, NS-MAN cannot generalize to unseen concept types and mostly fails in all cases with such queries. 

\begin{table}[t]
\vspace{-2.5mm}
    \centering

    \begin{tabular}{ccccc}
    \toprule
    \multirow{2}{3em}{\textbf{Method}} & \multicolumn{2}{c}{\textbf{SynHOTS-REF}}  & \multicolumn{2}{c}{\textbf{OCID-VLG}} \\
    & \textit{val} & \textit{test} & \textit{val} & \textit{test} \\
    \midrule
    gpt-4o \citep{GPT4} & 100.0 & 100.0 & 100.0 & 91.8 \\ 
    \midrule
    NS-VQA \citep{ns-vqa} & 99.0 & 36.2 & 99.2 & 9.6 \\ 
    NS-MAN (ours) & 99.8 & 87.1 & 99.2 & 52.8 \\
        \bottomrule
    \end{tabular}%
    \caption{\footnotesize
    Parsing accuracy (\%), measured as the percentage of correctly generated programs for the input query with ground-truth perception. The LLM transfers few-shot given 8 query-program examples from each dataset, while the below approaches are trained in the full provided data. The \textit{val} set contains seen concept types and vocabulary, while the \textit{test} set contains novel vocabulary for SynHOTS and novel vocabulary and concept types for OCID-VLG.}
    \label{tab:Tab7}
      \vspace{-1mm}
\end{table}

\noindent \textbf{Visual grounding} We compare the concept grounding capability of NS-MAN with foundation VLMs, such as CLIP \citep{CLIP} as well as \texttt{gpt-4o} \citep{GPT4} combined with set-of-mark (SoM) prompting from SAM \citep{SegmentA}, as in \citep{SetofMarkPU}. In the case of CLIP, we also use SAM to provide masks candidates, crop them and embed each one using CLIP to compute similarity with the language query. We note that both the validation and the test splits contain mixed queries that can contain category names, attributes or spatial relations. However, test split contains unseen vocabulary for SynHOTS and unseen category concepts for OCID-VLG. The spatial concepts are seen for both splits of both datasets. 
Results are given in Tab. \ref{tab:Tab8}.
We observe that in all cases of seen concept types, both splits of SynHOTS and validation split of OCID-VLG, NS-MAN outperforms the zero-shot methods. As expected, our grounder's performance in unseen concept types of OCID-VLG test degrades drastically (44.3 \%).
The margin is mostly due to spatial relation queries, which the SAM+CLIP baseline has no way to reason about. The SoM-GPT baseline has very strong performance in category and attribute grounding, but still struggles to perform multi-hop relational reasoning in an end-to-end fashion, just from the provided image and annotated markers. 
Finally, when we combine our semantic parser with the zero-shot VLM-based grounders, essentially just replacing our grounding modules with the VLM, we see a drastic boost in grounding accuracy.
We believe this result encourages the idea of using parsing as a proxy for referring expression reasoning, even in the case of generalist pretrained models, when applied to complex tabletop scenarios.

\begin{table}[t]
\vspace{-2.5mm}
    \centering

         \resizebox{\linewidth}{!}{%
    \begin{tabular}{cccccc}
    \toprule
    \multicolumn{2}{c}{\textbf{Method}} & \multicolumn{2}{c}{\textbf{SynHOTS-REF}}  & \multicolumn{2}{c}{\textbf{OCID-VLG}} \\
    \textit{Parser} & \textit{Grounder} & \textit{val} & \textit{test} & \textit{val} & \textit{test} \\
    \midrule
        \multicolumn{2}{c}{NS-MAN (Ours)} & 99.6 & 79.4 & 94.0 & 44.3 \\ 
    \midrule
    - & CLIP \citep{CLIP} + SAM \citep{SegmentA}  & 34.4 & 37.7 & 29.8 & 22.3 \\ 
    - & SoM-GPT \citep{SetofMarkPU} + SAM \citep{SegmentA} & 77.9 & 72.3 & 82.2 & 80.9 \\ 
    \midrule
    NS-MAN & CLIP \citep{CLIP} + SAM \citep{SegmentA} & 75.3 & 79.9 & 60.6 & 53.8 \\ 
    NS-MAN & SoM-GPT \citep{SetofMarkPU} + SAM \citep{SegmentA} & 92.5 & 90.9 & 87.1 & 85.8 \\  
        \bottomrule
    \end{tabular}%
    }
    \caption{\footnotesize
    Grounding accuracy (\%), measured as the percentage of correctly grounded referring expressions, measured as final masks with IoU $> 0.5$ with the ground-truth mask. Combining VLM-based grounders with semantic parser improves the reasoning capability of the VLMs in cases of spatial reasoning queries.}
    \label{tab:Tab8}
      \vspace{-1mm}
\end{table}

\textbf{Interactive Object Grasping} We compare our full NS-MAN framework with OWG \citep{TowardsOG}, a recent zero-shot work relying solely on pretrained models to produce a suitable grasp pose from a natural language instruction. This method is essentially an integration of the SoM-GPT + SAM grounding approach considered above with 
GR-ConvNet \citep{Antipodal} for grasp synthesis, which is the same underlying model we are using in NS-MAN. We also compare with CROG \citep{OCIDVLG}, a baseline end-to-end language-guided grasping model proposed with the OCID-VLG dataset.
We use the 4-DoF grasp annotations of OCID-VLG as ground-truth and report results in Tab. \ref{tab:Tab9} using the Jacquard metric J@1 as in the original work \citep{OCIDVLG}, which considers the IoU and relative angle between the predicted and ground-truth grasp rectangles.
Similar to grounding, we observe the same trends: Within distribution, NS-MAN provides the highest scores, with a delta of 7\% from OWG. When moving to unseen category concept types, NS-MAN's as well as the supervised baseline CROG's performance drop drastically, while OWG maintains high performance, as it relies solely on zero-shot VLM grounding.

\begin{table}[b]
\vspace{-2.5mm}
    \centering

    \begin{tabular}{ccc}
    \toprule
    \multirow{2}{3em}{\textbf{Method}} &  \multicolumn{2}{c}{\textbf{OCID-VLG}} \\
    & \textit{val} & \textit{test} \\
    \midrule
    CROG \citep{OCIDVLG}   & 77.2 & 42.1 \\ 
    OWG \citep{TowardsOG} &  75.8 & 72.6 \\ 
    NS-MAN (ours) & 82.8 & 42.3 \\
        \bottomrule
    \end{tabular}%
    \caption{\footnotesize
    Jacquard metric J@1 (\%), measured as the percentage of grasp predictions that have an IoU $>0.25$ and a relative angle within 30 degrees from ground-truth grasps. The \textit{val} set contains seen query vocabulary, while the \textit{test} set contains novel object queries.}
    \label{tab:Tab9}
      \vspace{-1mm}
\end{table}

Overall, we interpret the above results as a trade-off.
LLM/VLM-based methods offer significant benefits when considering completely unseen concepts.
NS-MAN achieves better / on-par results when in-distribution, or when considering unseen vocabulary of the same concept types, while doing so by being trained completely bottom-up from synthetic data.
On the one hand, if generalization is required, foundation model approaches are a favorable choice, with the additional costs of latency, cost, privacy and other factors. 
On the other hand, if efficiency-cost is required, and we assume that we have generated synthetic training data for all concept types that will be ever observed by the robot, our proposed framework serves as an effective lightweight alternative.
Further, as the results of this section suggest, LLMs for semantic parsing and/or VLMs for grounding can be drop-in replacements to corresponding modules of NS-MAN, thus boosting generalization to unseen concept types if in-the-wild scenarios are to be considered.

\section{Discussion}
\label{discussion}
In this section, we reflect on our results with regard to specific topics and discuss limitations and future work.

\subsection{Adapting to Novel Content}
One important benefit of the modular versus holistic design is the ability to adapt to novel content by only adapting the related module, instead of the entire pipeline \citep{tziafas}.
We believe that this translates to important benefits in terms of development cycles, as it alleviates the need for collecting large-scale multimodal data for training an end-to-end model.
In summary, the steps required for the proposed method to extend to novel concepts / tasks are:

\noindent \textbf{New concepts} require fine-tuning the VG(/SG) in a \textit{image-only} dataset annotated with the novel concepts and transferring the rest of the system without any further adaptation.
Even though from the experimental analysis of Sec.~\ref{exp-real} we conclude that a few examples per new concept are sufficient for visual adaptation, continuously incorporating new visual concepts would eventually outscale the capacity of the VG or lead to \textit{catastrophic forgetting}.
In the future, we plan to experiment with vision-language foundation models (e.g CLIP \citep{CLIP}) for zero-shot visual-language grounding.
Similarly, spatial concepts would require fine-tuning the SG networks for the new spatial concepts, without facing similar capacity issues due to the scarcity of spatial concepts used in referring expressions (with 11 concepts in this work we cover more than associated benchmarks, e.g. \citep{referit}). 

\noindent \textbf{New tasks} that involve new reasoning / control functionalities would require formally defining them as new primitives and integrating them in the DSL.
New task-related templates have to be generated to train the language parser, like we do in Sec.~\ref{exp-tasks}.
Even though our results suggest that the language parser can efficiently incorporate more tasks, the system is limited to the range of tasks that can be solved in a sequential fashion (chain of primitive steps), in order to be compatible with our DSL formalism.
Extending to more complex logic like conditionals and loops (e.g. \textit{``Keep the soda inside the bowl until you see a new item on the right"}) would require re-designing our language in an imperative rather than functional fashion.
Alternatively, one desirable future direction is to replace our supervised parser with a large language model \citep{gpt3} for zero-shot parsing of instructions to Pythonic code, akin to \citep{Socratic, CaP}.

\subsection{Handling Failure via Interactivity}
\begin{figure*}[!t]
    \centering
    \includegraphics[width=\textwidth]{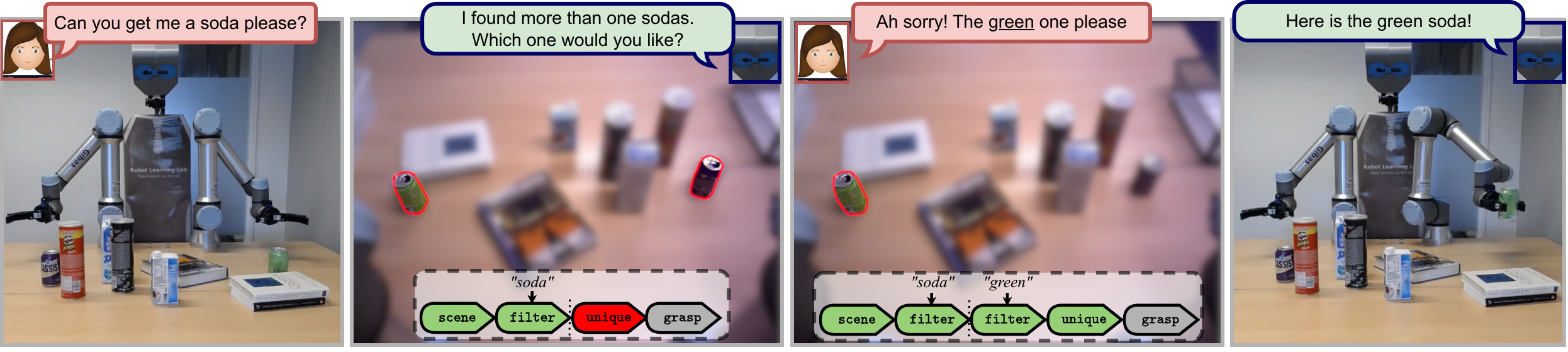}
    \caption{\footnotesize
    By type-checking and adding failure wrappers in all of our primitive implementations, the system is able to identify sources of failure and return a suitable response to the user. In this example, the original query \textit{(first)} is ill-posed as it refers to a soda object, while two sodas are present. This results in failure of the \texttt{unique} primitive, which will be prompted back to the user \textit{(second)}. The human responds with additional feedback \textit{(third)} which results in a correct final gasping behavior \textit{(fourth)}. Such failure handling behaviors allow our model to interact naturally with human users in a dialogue setting. }
    \label{fig:Fig9}
\end{figure*}
With this work we wish to highlight the practical benefits of interpretability in the context of human-robot interaction applications.
Beyond easiness of debugging and transparency of the model, this feature can augment models functionally by bringing humans in-the-loop.
For example, by adding suitable responses when a module fails at execution time, we can employ the system in an online dialogue setup, enabling the user to give feedback on failures caused by either ill-formed queries or other ambiguities in the scene (see Fig.~\ref{fig:Fig9}).
The \texttt{unique} primitive requires the input set to be unitary and therefore the execution will fail due to the presence of multiple matched objects.
The system raises a relevant template response back to the user and integrates their feedback to correct the generated program.
To achieve this, we process the feedback query to identify new present concepts from our concept memory.
When new concepts are identified, the parsed program is re-structured appropriately and the system re-runs execution.

\subsection{Training Time, Real-time Performance and Dynamic Environments}
Regarding training time, the entire curriculum training process discussed in Sec.~\ref{training} takes around 10 hours in a consumer GPU for the $500k$ scenes of \textit{SynHOTS-VQA}.
For inference, our end-to-end system (incl. the pretrained networks) can be used to produce a program at 4 fps in our hardware setup~\footnote{AMD Ryzen 7 3700X 8-core$*$16, NVIDIA GeForce RTX 3060}, with the main bottleneck being Mask R-CNN for localization.
In the future we plan to integrate high-efficiency detectors to increase our throughput.
Similarly to the previous subsection, failure handling in the implemented control primitives can be used to simulate closed-loop control, as in cases of dynamic environments the world state might change during execution.
A failure wrapper around the \texttt{grasp} primitive verifies that the target object state is the same as when the execution trace started (i.e., \texttt{scene} primitive) and otherwise re-runs the program with the updated state.

\subsection{Portability}
Besides sample-inefficiency, holistic approaches are limited to the agents / environments that were used to generate training data. 
In contrary, our approach disentangles the actual policy (represented as a program) from the perceptual and motor components (represented as functions in the program), and hence can be transferred to new agents / environment with minimal effort.
Similar to our experiments in Sec.~\ref{exp-real}, where we only adapt one module (VG) and transfer the overall system in a new visual domain, one could further replace the grasping module to use different arms or grippers and transfer to completely new robots and environments.

\section{Conclusion}
In this work we bring together deep learning techniques for perception, grasp synthesis and NLP with symbolic program synthesis and execution in an end-to-end hybrid system, aimed for interactive robot manipulation applications. 
We design a dedicated language that implements visuospatial reasoning as primitive operations. We exploit linguistic cues in the input instruction to synthesize a program composed of such primitives. 
Programs interface with visual/spatial grounding and grasp modules to ground concepts and control the robot respectively.
We generate a synthetic tabletop dataset with rich scene graph and language-program annotations, paired with a real RGB-D scenes dataset, which we make publicly available.
Extensive evaluation through a VQA task showcases that our method achieves near-perfect accuracy in-domain, while being fully interpretable and sample-efficient compared to baselines.
Generalization experiments show that the vocabulary-agnostic formulation of our language and model enables better generalization to unseen concept words compared to previous works. 
Also, we show that with our modular design, the system can transfer to natural scenes with few-shot adaptation of the visual grounder, as well as transfer to more manipulation tasks with few-shot adaptation of the language parser module. 
We integrate our model with a robot framework and perform experiments for an interactive object picking task, both in simulation and with a real robot. Robot experiments demonstrate high success rate, and robustness to user instructions, with interpretability leveraged to actively detect reasoning failures and inform the user.

\bibliographystyle{SageH}


\bibliography{main}  

\appendix

\begin{figure*}[!t]
    \includegraphics[width=1\textwidth]{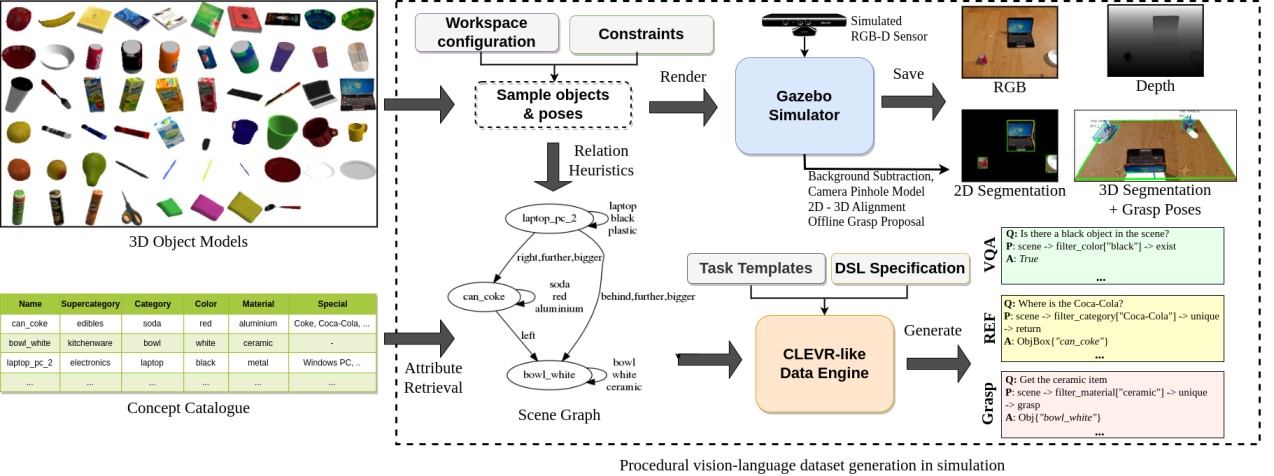} 
\caption{\footnotesize
Dataset generation pipeline. Given a catalogue of 3D object models and a table of domain-specific annotations, the pipeline samples random initial world states according to a workspace specification and given constraints (e.g. number of objects, the inclusion of distractors, etc.) and renders image data with a simulated depth sensor in Gazebo. The world state is parsed in a symbolic scene graph containing all attribute and relation information. We process the visual data offline to generate 2D and 3D segmentation masks / boxes and run a grasp synthesis network to propose optimal grasp-poses per object. A CLEVR-like data engine tailored for our domain and tasks samples concepts from the scene graph to generate query-program-answer triplets for REF/VQA/Grasping tasks.
}
\label{fig:Fig10}
\end{figure*}

\section{Dataset Details}
\label{appendixData}
We present the synthetic dataset generation pipeline in Fig.~\ref{fig:Fig10}. 
It is noted that this dataset also includes REF / Grasping tasks by rephrasing referring expressions as questions with the location of the target object as the final answer.
A comparison of different statistics / features of SynHOTS with other popular VQA/EQA datasets are given in Table~\ref{tab:Tab10}.
The detailed object and concept catalogue included in both versions of our dataset is given in Table~\ref{tab:Tab11}, while the vocabulary used to represent the concepts, as well as the held-out vocabulary for the generalization-test split is given in Table~\ref{tab:Tab12}.
The task templates used to generate text-program data are presented in Table~\ref{tab:Tab13}.

\begin{table*}[!t]
    \centering
    \resizebox{\textwidth}{!}{\begin{tabular}{lccccccc}
    \toprule
    \multirow{2}{3em}{\textbf{Dataset}} &
    \multirow{2}{4em}{\textbf{Synthetic}} &
    \textbf{Vision} &
    \textbf{Num.} &
    \textbf{Num.Obj.} &
    \textbf{Num.} &
    \multirow{2}{3em}{\textbf{Parses}} &
    \multirow{2}{4em}{\textbf{Grasp. Annot.}} \\
     &
     &
     \textbf{Data} &
     \textbf{Questions} &
     \textbf{Categories} &
     \textbf{Scenes} &
     &
     \\
     \midrule
     \multicolumn{8}{l}{\textbf{VQA/EQA Datasets}} \\
     VQA‑2.0 \citep{VQA2}  & \textcolor{red}{\XSolidBrush} & RGB & 1.1M & 80 (MS COCO) \citep{ms-coco}  & 265K & \textcolor{red}{\XSolidBrush} & \textcolor{red}{\XSolidBrush} \\
     AOK‑VQA \citep{AOKVQA} &  \textcolor{red}{\XSolidBrush} & RGB & 25K &  80 (MS COCO) \citep{ms-coco} & 205K & \textcolor{red}{\XSolidBrush} & \textcolor{red}{\XSolidBrush} \\
     EQA \citep{eqa}  & \textcolor{green}{\CheckmarkBold} & 3D & 9K &  80 (SUNCG) \citep{SUNSG} & 774 & \textcolor{red}{\XSolidBrush} & \textcolor{red}{\XSolidBrush} \\
     \midrule
     \multicolumn{8}{l}{\textbf{Neurosymbolic VQA Datasets}} \\
     CLEVR &  \textcolor{green}{\CheckmarkBold} & RGB & 850K & - (primitives) & 100K &  \textcolor{green}{\CheckmarkBold} & \textcolor{red}{\XSolidBrush} \\
     SynHOTS (ours) &  \textcolor{green}{\CheckmarkBold} & 3D & 600K & 58 & 9.6K &  \textcolor{green}{\CheckmarkBold} &  \textcolor{green}{\CheckmarkBold} \\
     \bottomrule
    \end{tabular}}%
    \caption{\footnotesize
    Comparison of features and statistics between various VQA/EQA datasets and SynHOTS. Our proposed dataset is the only one who combines grasp annotations with question-answering in an end-to-end fashion, together with symbolic parses of both the vision and the question modality (programs).}
    \label{tab:Tab10}
\end{table*}

\begin{table}[!ht]
  \caption{
  Attribute concept catalogue of SynHOTS (\cblue{blue}) and HOTS (\cred{red}) datasets. The number in color represents the total number of object instances of a concept.}
  \label{tab:Tab11}
  \centering
  \resizebox{0.9\linewidth}{!}{
  \begin{tabular}{lcr}
    \toprule
    \textbf{Annotation} & \textbf{Number} & \textbf{Classes} \\
    \midrule
      &  & edibles(\cblue{13},\cred{13}), electronics(\cblue{4},\cred{5}), \\
    Supercategory & \cblue{5},\cred{5} &  fruits(\cblue{6},\cred{6}), kitchenware(\cblue{18},\cred{11}),  \\
      & & stationery(\cblue{17},\cred{11}) \\
\midrule
     &  & apple(\cblue{1},\cred{1}), banana(\cblue{1},\cred{1}), book(\cblue{6},\cred{3}),\\
          &  &  bowl(\cblue{4},\cred{1}), soda(\cblue{5},\cred{5}), cup(\cblue{4},\cred{3}),\\
           & & fork(\cblue{1},\cred{2}), juice(\cblue{4},\cred{3}), keyboard(\cblue{1},\cred{1}),\\
            & & knife(\cblue{1},\cred{1}), laptop(\cblue{2},\cred{1}), lemon(\cblue{1},\cred{1}),\\
         Category   &  \cblue{25},\cred{25} & marker(\cblue{3},\cred{2}), milk(\cblue{1},\cred{2}), stapler(\cblue{0},\cred{1}),\\
            & & mouse(\cblue{1},\cred{2}), orange(\cblue{1},\cred{1}), peach(\cblue{1},\cred{1})\\
            & & mug(\cblue{4},\cred{0}), pear(\cblue{1},\cred{1}), pen(\cblue{4},\cred{3}),\\
            & & plate(\cblue{3},\cred{2}), Pringles(\cblue{3},\cred{3}), scissors(\cblue{1},\cred{2}),\\
            & & sponge(\cblue{3},\cred{0}), spoon(\cblue{1},\cred{2}), monitor(\cblue{0},\cred{1}) \\
\midrule
          &  & red(\cblue{9},\cred{6}), yellow(\cblue{6},\cred{4}),\\
     &  &  purple(\cblue{2},\cred{2}), pink(\cblue{3},\cred{1})\\
     Color &  \cblue{10},\cred{10} &  black(\cblue{7},\cred{9}), silver(\cblue{3},\cred{6}),\\
     &  &  orange(\cblue{4},\cred{2}), green(\cblue{9},\cred{4}),\\
     &  &  blue(\cblue{8},\cred{5}), white(\cblue{7},\cred{6})\\
\midrule
          &  & glass(\cblue{3},\cred{1}), metal(\cblue{4},\cred{8}), paper(\cblue{7},\cred{7}),\\
     Material & \cblue{8},\cred{7} &  ceramic(\cblue{11},\cred{5}), aluminium(\cblue{5},\cred{5}),\\
      &   &  organic(\cblue{6},\cred{6}), plastic(\cblue{19},\cred{14}), synthetic(\cblue{3},\cred{0})\\
          \bottomrule
  \end{tabular}}
\end{table}

\begin{table}[!ht]
  \caption{\footnotesize
  Concept words and synonyms included in the vocabulary of the training data. Samples that contain words in red are held out in the generalization-test split used in our experiments.}
  \label{tab:Tab12}
  \centering
  \resizebox{0.9\linewidth}{!}{
  \begin{tabular}{lr}
    \toprule
    \textbf{Concept} & \textbf{Vocabulary}  \\
    \midrule
      & apple,banana,book,bowl/food bowl,cup,fork,knife \\ 
      & juice box/drink/\cred{package},keyboard,lemon, \\
      Category & laptop/computer/\cred{PC}/\cred{computer screen},milk drink/\cred{box}, \\ 
       & mouse,mug/\cred{coffee cup},orange,peach,pear,pen,marker, \\ 
    &  plate,Pringles box/\cred{potato chips package/product},scissors, \\ 
    & soda/soda drink/\cred{soft drink/product/can},sponge,spoon \\ 
    \midrule
     & red,yellow,black,blue/\cred{cyan}, \\
     Color & orange,green,purple/\cred{magenta}, \\ 
     & pink,white,silver/\cred{gray}  \\ 
    \midrule
     & glass/\cred{transparent},paper,organic,metal/\cred{metallic} \\
     Material & synthetic/\cred{polymer},aluminium/steel/\cred{tin}, \\ 
     & ceramic/\cred{porcelain},plastic/\cred{consumable} \\ 
    \midrule
    & Coca-Cola/\cred{Coke}/\cred{Cola},Coca-Cola Zero/\cred{Coke Zero},\cred{Cola Zero} \\
    & Mac laptop/computer/\cred{Windows laptop/computer},beer cup/\cred{hexagonal cup} \\
     & coffee cup/\cred{tall cup}/\cred{tumbler},Sci-Fi book,animals book/\cred{birds book}, \\ 
    Open & coding book/\cred{software design book/textbook},Computer Vision book/\cred{textbook}, \\ 
    & mystery novel/\cred{Sherlock Holmes book},RIPE book/\cred{self-help book}, \\
    & Fanta,Pepsi,Sprite,apple/mango/lemon/\cred{citrus,cranberry,strawberry} juice, \\
    & Original/\cred{Sour Cream/Hot \& Spicy} Pringles \\
    \bottomrule
  \end{tabular}}
\end{table}

\begin{table}[!ht]
  \caption{\footnotesize
  Catalogue of task template families used to generate \textit{SynHOTS}, their associated tasks, number of total sub-templates per family, and a given example, where letters in [,] correspond to concepts sampled from each scene graph to generate the query (Y: category, M: material, C: color, R: relation, L: location, H: hyper-relation, X: open instance-level category).}
  \label{tab:Tab13}
  \centering
  \resizebox{0.9\linewidth}{!}{
  \begin{tabular}{cccc}
    \toprule
    \textbf{Template}  & \textbf{Task} & \textbf{Num.Sub-} & \multirow{2}{3em}{\textbf{Example}}  \\
      \textbf{Family} & \textbf{\textbf{Types}} & \textbf{Templates} & \\
    \midrule 
    \multirow{2}{6em}{compare integer} & \multirow{2}{10em}{VQA\{compare number\}} & \multirow{2}{1em}{33} & \footnotesize{`Are there fewer [C] [M] [Y]s} \\
    & & & \footnotesize{than [C2] [M2] [Y2]s?'} \\
        \midrule
    \multirow{2}{6em}{comparison} & \multirow{2}{10em}{VQA\{compare attribute\}} & \multirow{2}{1em}{60} & \footnotesize{`Do the [L] [C] [M] [Y] and the [C2]} \\
    & & & \footnotesize{[M2] [Y2] have the same material?'} \\
        \midrule
            \multirow{2}{6em}{zero\_hop} & VQA\{count,query,exist\} & \multirow{2}{1em}{8} & \footnotesize{`The [L] [M] [Y] has what color?'} \\
    & REF, Grasp & & \\    
        \midrule
            \multirow{2}{6em}{one\_hop} & VQA\{count,query,exist\} & \multirow{2}{1em}{15} & \footnotesize{`There is a [C] [M] [Y]; is there a [X] [R] it?'} \\
    & REF, Grasp & & \\    
        \midrule
            \multirow{2}{6em}{two\_hop} & VQA\{count,query,exist\} & \multirow{2}{1em}{15} & \footnotesize{`What is the [C3] [M3] [Y3] [that is] [R2] the} \\
    & REF, Grasp & &  \footnotesize{[C2] [M2] [Y2] [that is] [R] the [C] [M] [Y]?'} \\  
    \midrule
    \multirow{2}{6em}{hyper\_one\_hop} & VQA\{count,query,exist\} & \multirow{2}{1em}{34} & \footnotesize{`There is a [C2] [M2] thing [that is] [H] the [L]} \\
    & REF, Grasp & & \footnotesize{[C] [M] [Y] than the [X]; what material is it?'} \\
            \midrule
    \multirow{2}{6em}{hyper\_two\_hop} & VQA\{count,query,exist\} & \multirow{2}{1em}{10} & \footnotesize{`How many [C4] [M4] [Y4]s are [H] the [C] [M] [Y] than} \\
    & REF, Grasp & & \footnotesize{the [C3] [M3] [Y3] [that is] [R] the [C2] [M2] [Y2]?'} \\    
           \midrule
    \multirow{2}{6em}{single\_and} & VQA\{count,query,exist\} & \multirow{2}{1em}{24} & \footnotesize{`Where is the [C3] [M3] [Y3] that is [both]} \\
    & REF, Grasp & & \footnotesize{[R2] the [L2] [Y2] and [R] the [X]?'} \\     
               \midrule
    \multirow{2}{6em}{single\_or} & VQA\{count,query,exist\} & \multirow{2}{1em}{24} & \footnotesize{`How many objects are [either] [C] [M] or [C3]} \\
    & REF, Grasp & & \footnotesize{[M3] [that are] [R] the [L2] [C2] [M2] [Y2]?'} \\    
               \midrule
    \multirow{2}{6em}{same\_relate} & VQA\{count,query,exist\} & \multirow{2}{1em}{54} & \footnotesize{`Grasp the [L2] [M2] object that} \\
    & REF, Grasp & & \footnotesize{has the same color as the [X]'} \\    
               \midrule
    \multirow{2}{6em}{return} & \multirow{2}{5em}{REF, Grasp} & \multirow{2}{1em}{18} & \footnotesize{`Grab the [L] [C] [M] [Y]'} \\
    &  & & \\    
                   \bottomrule
  \end{tabular}}
\end{table}

\section{Spatial Relation Resolution}
\label{appendixSpatial}
\begin{figure}[!b]
\begin{tabular}{c c}
    \includegraphics[width=0.47\textwidth]{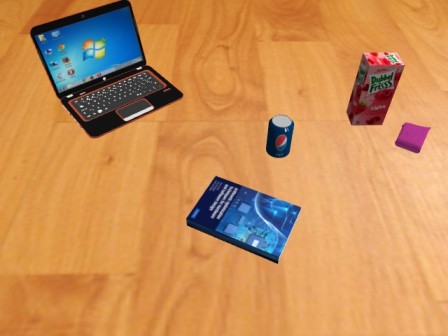} &  \hspace{-2mm}\includegraphics[width=0.47\textwidth]{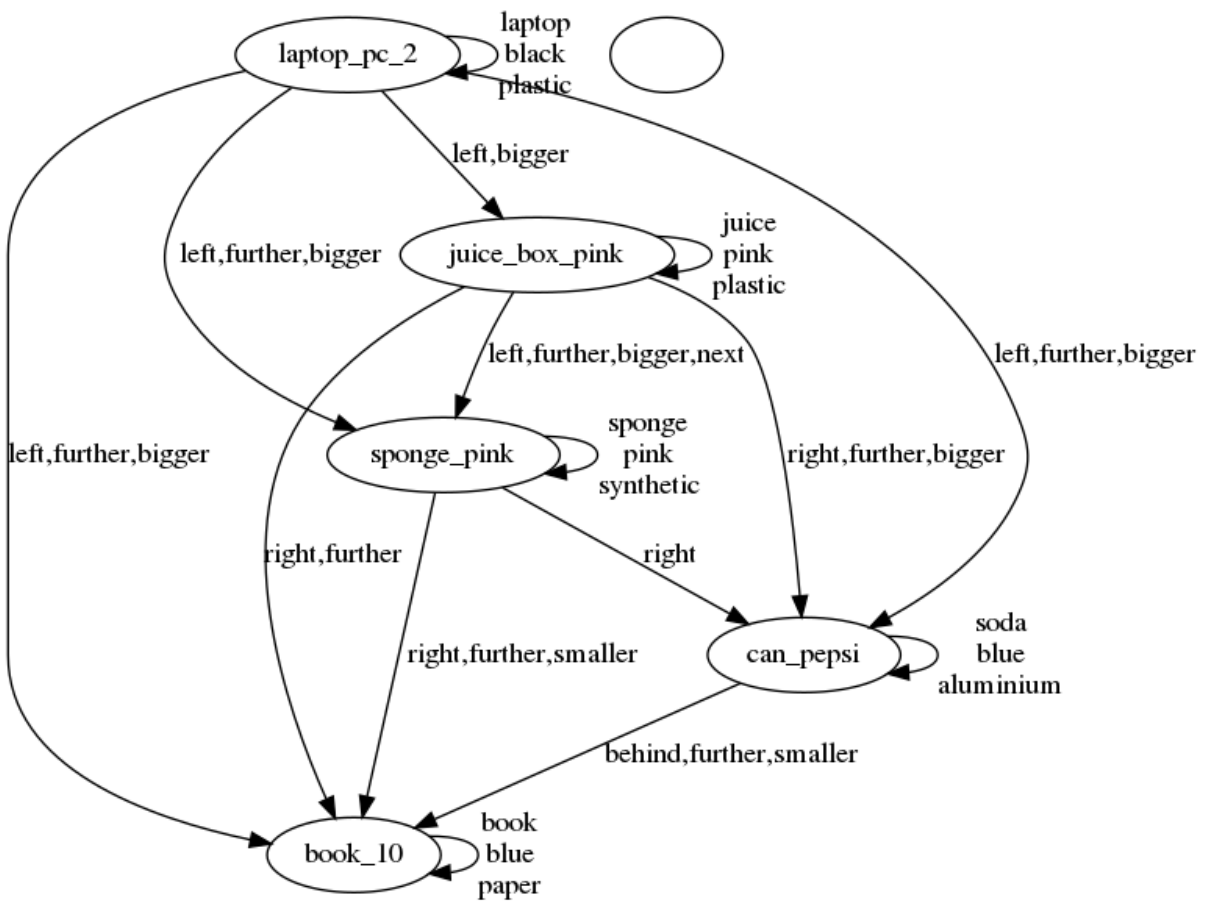}\\
    \multicolumn{2}{c}{\includegraphics[width=1\textwidth, trim= 4cm 14.5cm 1.5cm 2cm,clip=true]{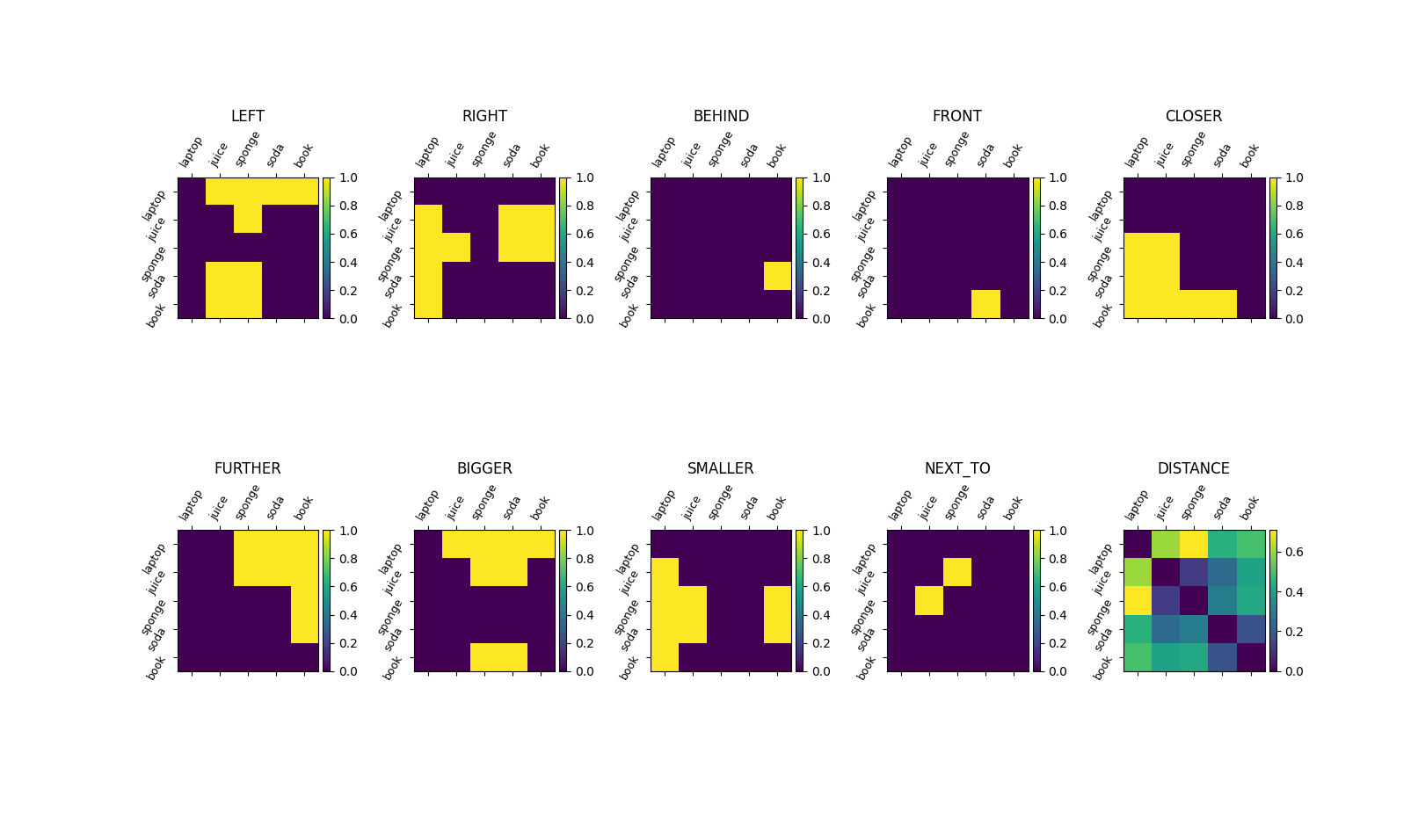}}\\
    \multicolumn{2}{c}{\includegraphics[width=1\textwidth, trim= 4cm 4.5cm 1.5cm 13cm,clip=true]{imgs/sceneGraph_rels.png}}
\end{tabular}
\caption{\footnotesize
Example image rendered in Gazebo environment \textit{(top-left)}, extracted scene graph with dense attribute and relation annotations \textit{(top-right)}, pair-wise spatial relations labels for all depicted objects \textit{(bottom)}. The pair-wise maps are used as supervision to train the spatial grounder networks.%
}
\label{fig:Fig11}
\end{figure}

In this section, we describe the heuristic functions $\zeta_r$ used to produce annotations for spatial relations $r$ included in our tabletop domain.
The vocabulary set of spatial concepts considered is $\mathcal{R} = $ \{\textit{``left", ``right", ``behind", ``front", ``closer", ``further", ``bigger", ``smaller", ``next to"}\}.
The choice of vocabulary takes into account elementary spatial concepts that are often used to disambiguate same object instances from one another in natural language. 
See Fig.~\ref{fig:Fig11} for an illustration of a parsed scene graph with detailed pairwise spatial relations in a synthetic scene. 
For a scene graph $\mathcal{G} = \left \{ \mathcal{V}, \mathcal{E}, \mathcal{X_V}, \mathcal{X_E} \right \}$ and any two objects with unique indices $n,m \in \mathcal{V}$, binary relation features $\zeta_r(n,m)  \in \{0,1\}$ are computed for each pair $(n,m) \in \mathcal{E}$ and value of $r \in \mathcal{R}$ by:

\begin{equation*}
    \zeta_{\textit{``left"}}(n,m) = \left [ x_n + \frac{l_n^x}{2}  < x_m - \frac{l_m^x}{2}  \right ]
\end{equation*}

\begin{equation*}
    \zeta_{\textit{``right"}}(n,m) = \left [ x_n - \frac{l_n^x}{2}  > x_m + \frac{l_m^x}{2}  \right ]
\end{equation*}

\begin{equation*}
    \zeta_{\textit{``behind"}}(n,m) = \left [\left | x_n - x_m \right | < \frac{l^x_n + l^x_m}{2}  \right ] \cdot \left [ \Delta y_{nm} > \frac{\Delta l^y_{mn}}{2} \right ]
\end{equation*}

\begin{equation*}
    \zeta_{\textit{``front"}}(n,m) = \left [\left | x_n - x_m \right | < \frac{l^x_n + l^x_m}{2}  \right ] \cdot \left [ \Delta y_{nm} < - \frac{\Delta l^y_{mn}}{2} \right ]
\end{equation*}

\begin{equation*}
    \zeta_{\textit{``closer"}}(n,m) = \left [ y_n + \frac{l^y_n}{2} < y_m - \frac{l^y_m}{2} \right ]
\end{equation*}

\begin{equation*}
    \zeta_{\textit{``further"}}(n,m) = \left [ y_n - \frac{l^y_n}{2} > y_m + \frac{l^y_m}{2} \right ]
\end{equation*}

\begin{equation*}
    \zeta_{\textit{``bigger"}}(n,m) = \left [ l^x_n \cdot l^y_n \cdot l^z_n > l^x_m \cdot l^y_m \cdot l^z_m + \Delta_{size\_thr} \right ]
\end{equation*}

\begin{equation*}
    \zeta_{\textit{``smaller"}}(n,m) =  \left [ l^x_n \cdot l^y_n \cdot l^z_n < l^x_m \cdot l^y_m \cdot l^z_m - \Delta_{size\_thr} \right ]
\end{equation*}

\begin{equation*}
    \zeta_{\textit{``next"}}(n,m) = \left [ \left \| \mathbf{p}_n - \mathbf{p}_m \right \|^2 \leq \Delta_{next\_thr} \right ]
\end{equation*}

\noindent where $\mathbf{p}_n = (x_n, y_n, z_n)^T$ denotes the centroid and $(l_n^x, l_n^y, l_n^z)$ the dimensions of the approximate 3D bounding box of object $n$, normalized according to the workspace dimensions.
The $[\cdot]$ operator denotes evaluating the input condition for true/false. 
We empirically select thresholds $\Delta_{size\_thr}=0.45$ and $\Delta_{next\_thr}=0.25$ for resolving size and promiximity relations.

We extend basic relations with higher-order (\textit{hyper-relations}), which are used to resolve queries such as: \textit{``The bowl that is closer to the coca-cola than the cereal box"}. Such queries require considering the relative relation between a source $n$ and two target objects $m,k$ and thus are treated as a separate primitive in our library.
We implemented two distance-based hyper-relation concepts, namely: $\mathcal{H} = $ \{\textit{``closer to/than", ``further from/than"}\}, whose heuristics are given by:
\begin{equation*}
  \zeta(n,m,k)_{\textit{``closer"}} = \left [ \left \| \mathbf{p}_n - \mathbf{p}_m  \right \|^2 - \left \| \mathbf{p}_n - \mathbf{p}_k  \right \|^2 <0 \right ]
\end{equation*}
\begin{equation*}
\zeta(n,m,k)_{\textit{``further"}} = \left [ \left \| \mathbf{p}_n - \mathbf{p}_m  \right \|^2 - \left \| \mathbf{p}_n - \mathbf{p}_k  \right \|^2 >0 \right ]
\end{equation*}

We highlight that the coordinates for each object are expressed with respect to the robot base frame (which is aligned with the bottom middle of the tabletop), so spatial relations are resolved according to the robot's perspective.
We expect that the user queries the robot having this convention in mind. 
In the future, we plan to add human tracking to our system, allowing us to transform coordinates on-the-fly and resolve spatial relations with respect to arbitrary perspectives.

\section{Illustrations of Program Execution}
\label{appendixExamples}
\begin{figure*}[!ht]
\begin{tabular}{c c}
    \includegraphics[width=1\linewidth]{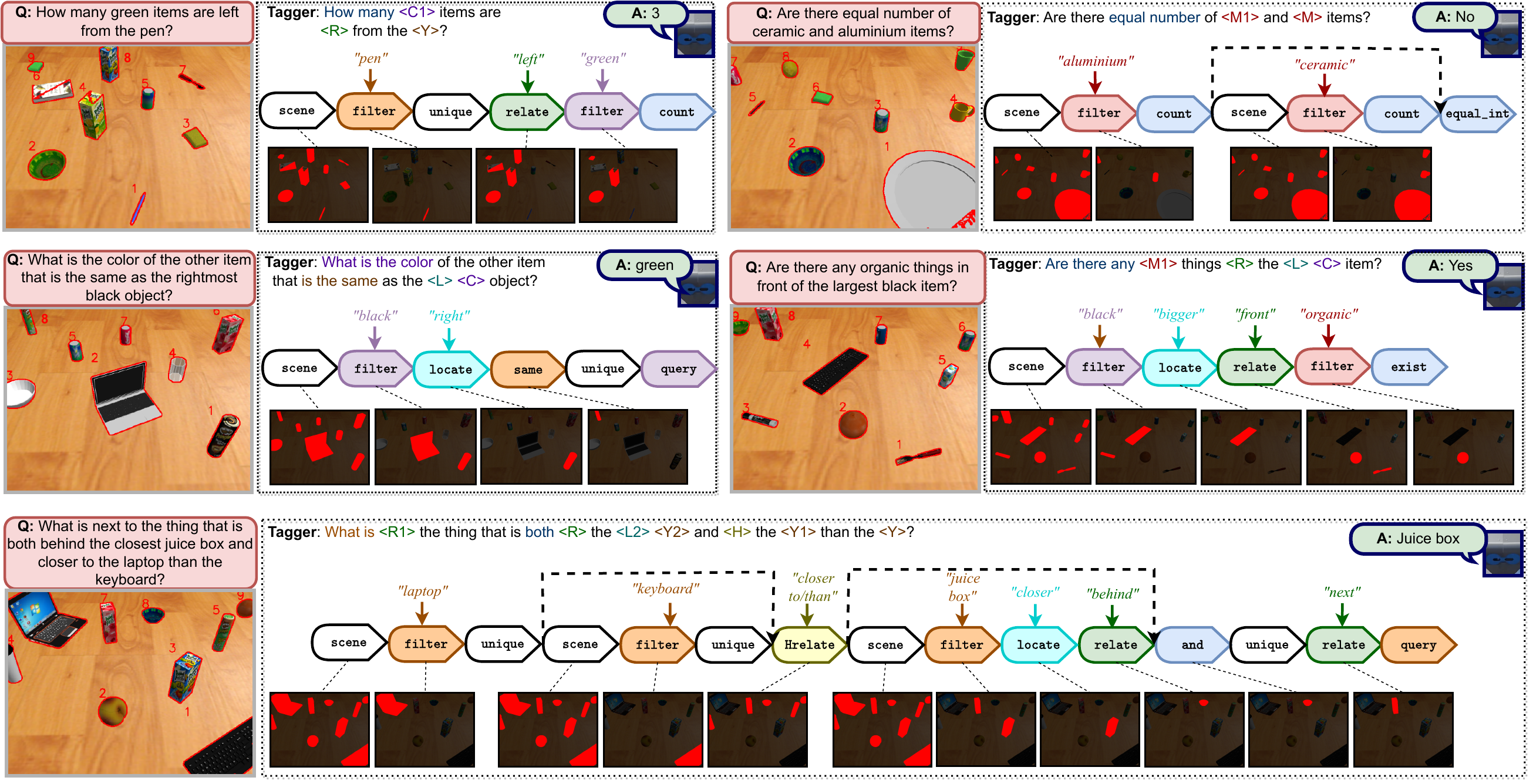} \\
\end{tabular}
\caption{\footnotesize
Illustration of VQA execution traces in scenes of our synthetic dataset. Execution steps that output object states are visualized as segmentation masks over the input RGB image, using the localization results. Concept arguments, i.e. category, color, and material are color-coded with brown, purple, and red respectively, while relations, locations, and hyper-relations with green, emerald, and yellow, and symbolic primitives with potential integer arguments are color-coded in blue.}%
\label{fig:Fig12}
\end{figure*}

\begin{figure*}[!t]
\begin{tabular}{c c}
    \includegraphics[width=1\linewidth]{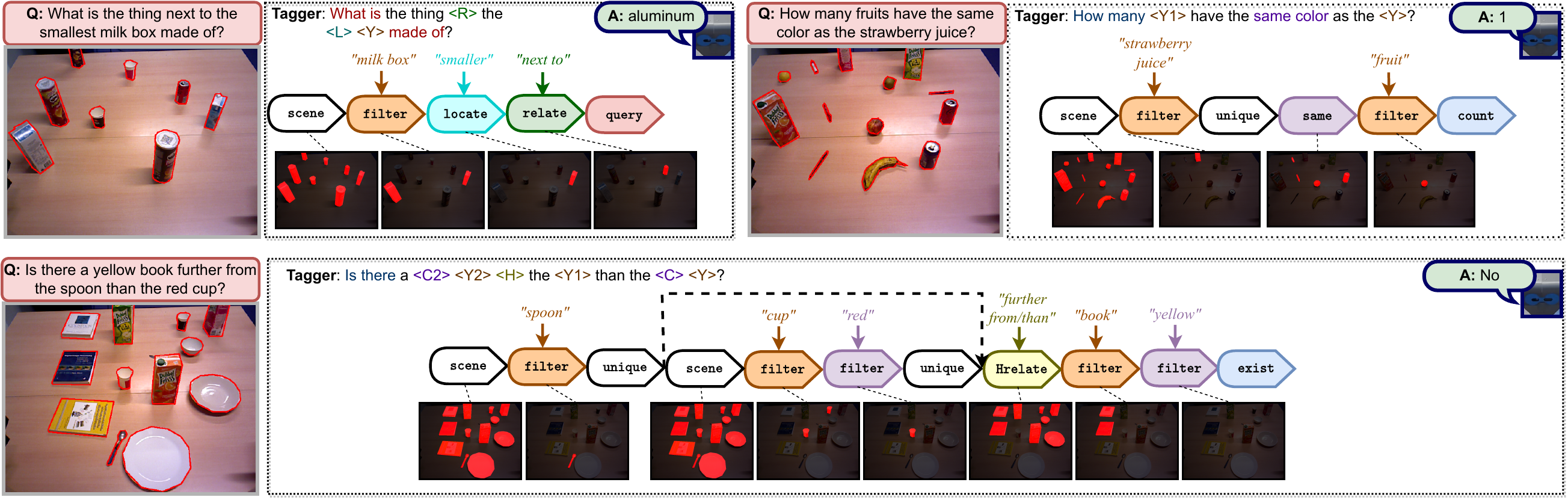} \\
    \includegraphics[width=1\textwidth]{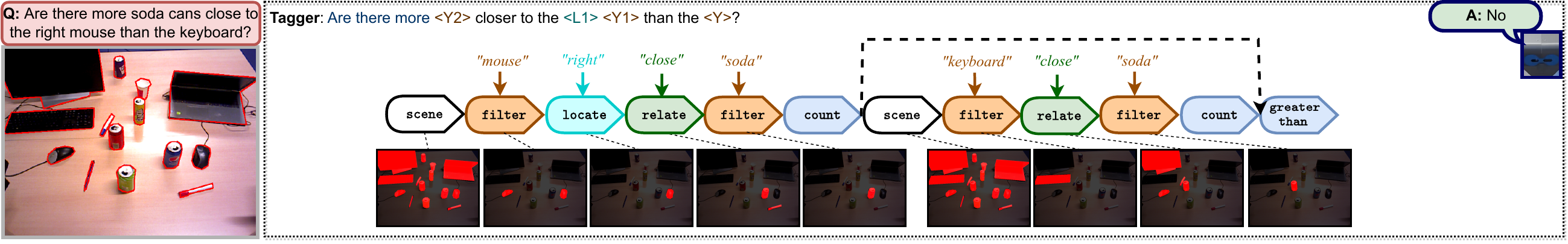}
\end{tabular}
\caption{\footnotesize
\footnotesize
Illustration of VQA execution traces in scenes of the released HOTS dataset. Execution steps that output object states are visualized as segmentation masks over the input RGB image, using the localization results. Concept arguments, i.e. category, color, and material are color-coded with brown, purple, and red respectively, while relations, locations, and hyper-relations with green, emerald, and yellow, and symbolic primitives with potential integer arguments are color-coded in blue. }%
\label{fig:Fig13}
\end{figure*}

In Fig.~\ref{fig:Fig12} and Fig.~\ref{fig:Fig13} we present running examples for different synthetic scenes in both versions of our dataset, covering a variety of compositional capabilities of our implemented model. 

\end{document}